%% file: main.tex
\documentclass[10pt,twocolumn,letterpaper]{article}

\def\confName{WACV}
\def\confYear{2027}
\def\wacvPaperID{2180} 

\usepackage[final, algorithms]{wacv}

\usepackage{algorithm}
\usepackage{algpseudocode}
\usepackage[accsupp]{axessibility} 
\usepackage{orcidlink}
\usepackage{amsthm}
\usepackage{multirow}

\newtheorem{proposition}{Proposition}
\definecolor{linkblue}{rgb}{0.21,0.49,0.74}
\AtBeginDocument{\hypersetup{colorlinks=true,allcolors=linkblue}}

\title{Data-free Multistep Diffusion Distillation}
\author{Ngoc Duy Tran$^{1}$ \quad Kien Do$^{2}$ \quad Duc Thanh Nguyen$^{3}$ \quad Truyen Tran$^{1}$\\
{\normalsize $^{1}$Deakin Applied Artificial Intelligence Initiative, Deakin University, Australia}\\
{\normalsize $^{2}$VinSmart Future, Vietnam \quad $^{3}$School of Information Technology, Deakin University, Australia}\\
{\small\tt s22414149@deakin.edu.au}}

\begin{document}

\maketitle

\newcommand{\Expect}{\mathbb{E}}
\newcommand{\Real}{\mathbb{R}}
\newcommand{\Var}{\mathbb{V}}
\newcommand{\Data}{\mathcal{D}}
\newcommand{\Loss}{\mathcal{L}}
\newcommand{\Normal}{\mathcal{N}}
\newcommand{\Id}{\mathrm{I}}
\newcommand{\Rectify}{\mathrm{Rectify}}
\newcommand{\ELBO}{\text{ELBO}}
\newcommand{\diag}{\text{diag}}
\newcommand{\softmax}{\text{softmax}}
\newcommand{\argmin}[1]{\underset{#1}{\text{argmin}}}
\newcommand{\argmax}[1]{\underset{#1}{\text{argmax}}}
\newcommand{\betamax}{\overline{\beta}_{\text{max}}}
\newcommand{\betamin}{\overline{\beta}_{\text{min}}}
\newcommand{\Model}{\text{PDMD}}

\begin{figure*}[t]
\centering
\includegraphics[width=\linewidth]{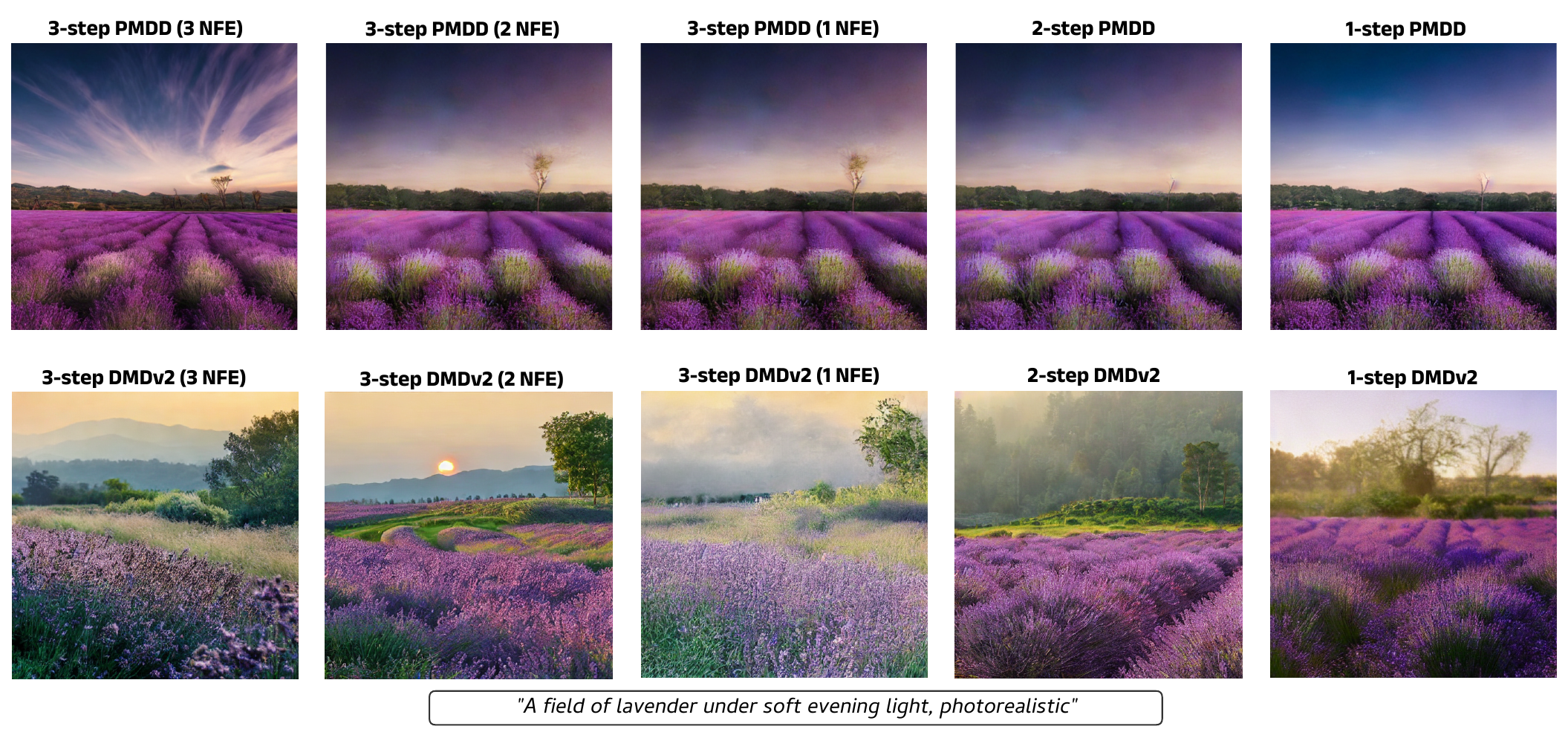}
\caption{Generations of PMDD and data-free DMDv2 across different inference budgets (NFE: neural function evaluations). DMDv2 shows strong inconsistency across inference budgets and between previously distilled 1-step and 2-step models. In contrast, PMDD maintains consistent structure and high-quality outputs across all sampling budgets using a single model.
\label{fig:teaser}}
\end{figure*}

\begin{abstract}
\input{sections/abstract}
\end{abstract}

\section{Introduction}
\input{sections/intro}

\section{Preliminary}
\label{sec:Preliminary}
\input{sections/prelim}

\section{Method}
\label{sec:Method}
\input{sections/method}

\section{Experiment}
\label{sec:Experiment}
\input{sections/exp}

\section{Related Work}
\input{sections/relate}

\section{Conclusion}
\input{sections/discuss}

\appendix
\tableofcontents
\input{sections/appendix}

{\small
\bibliographystyle{splncs04}
\bibliography{reference}
}

\end{document}

%% file: sections/abstract.tex
One-step diffusion distillation achieves strong performance on small datasets but remains insufficient for high-resolution or complex text-to-image settings. Extending to multistep distillation often requires training data, suffers from early-step degradation, inconsistent outputs across sampling budgets, and incurs high computational cost. We identify that these limitations stem from simultaneous training, where a single adapted network approximates multiple score functions of different student distributions at once, causing biased distribution matching and unstable optimization. Therefore, we propose Progressive Multistep Data-free Distillation (PMDD), which assigns a dedicated adapted network per student distribution while avoiding the memory overhead of maintaining multiple adapted networks. Combined with a stage-wise unforget loss to preserve earlier-step generations, PMDD reduces optimization bias in the student, provides faster convergence, and a single unified model that supports multiple inference budgets. Extensive experiments show that PMDD matches or outperforms data-free few-step baselines at every multistep budget, reaches FID 1.99 on ImageNet 64$\times$64 in two steps, below its 512-NFE teacher (2.32), and achieves FID 7.76 on zero-shot COCO 512$\times$512, while requiring 4--5$\times$ less training compute than data-free DMDv2 to support three inference budgets.

%% file: sections/intro.tex
Diffusion models have attained significant success in generative tasks, demonstrating state-of-the-art performance in a wide range of domains such as image generation~\citep{DBLP:conf/uai/SongGSE19,song2020score,ho2020denoising,song2020denoising}, audio synthesis~\citep{DBLP:conf/iclr/ChenZZWNC21,DBLP:conf/iclr/KongPHZC21}, and text generation~\citep{DBLP:conf/nips/AustinJHTB21,DBLP:conf/nips/GulrajaniH23,DBLP:conf/icml/LouME24}. Diffusion-based image generation models adopt an iterative denoising process which gradually removes noise from a noisy intermediate sample to reconstruct a high-quality image. However, the sampling process is inherently slow, typically requiring hundreds or thousands of neural function evaluations (NFEs), which makes the models expensive for many real-world applications.

To overcome this limitation, recent research has explored distilling a (diffusion) teacher into a (diffusion) student. These approaches fall into two families. $\emph{Trajectory-matching}$ methods~\citep{luhman2021knowledge,salimans2022progressive,song2023consistency,luo2023latent,DBLP:journals/corr/abs-2412-16906,DBLP:conf/iclr/KimLLMTUHME24} regress the student onto the teacher's deterministic ODE trajectory, and typically require real samples or adversarial training to match their teachers. $\emph{Distribution-matching}$ methods, motivated by frameworks such as GMMNs~\citep{DBLP:conf/icml/LiSZ15} and GANs~\citep{goodfellow2014generative}, instead learn a one-step mapping $x_{0|\phi,T}=g_{\phi}\left(x_{T},T\right)$ from Gaussian noise $x_{T}\sim\mathcal{N}\left(0,\mathcal{I}\right)$ to clean data, and align the $\emph{distribution}$ of student outputs with that of the teacher. An adapted network $\varphi$ is introduced to estimate the score function of the student distribution and guide it toward the teacher distribution.

While effective on small- and medium-scale datasets such as CIFAR-10 and ImageNet $64\times64$, one-step distillation struggles with complex text-image compositions and high-resolution synthesis~\citep{DBLP:journals/corr/abs-2410-20898,nguyen2024swiftbrush,yin2024one,yin2024improved,zhou2024score}. Therefore, extending distillation to few steps is necessary to achieve higher fidelity generation while preserving most of the sampling speedup.

To enable multistep sampling, prior approaches naively extend the one-step framework, allocating a single adapted network $\varphi$ to \textbf{simultaneously} approximate $N$ score functions corresponding to distributions produced by $g_{\phi}\left(x_{t_{i}},t_{i}\right)$ ($i=1,...,N$)~\cite{DBLP:journals/corr/abs-2505-12674,yin2024improved,DBLP:journals/corr/abs-2406-04103}. This single design choice is, we argue, the root cause of the field's current limitations. Forcing one network to fit $N$ heterogeneous score targets within a shared parameter space induces $\emph{negative transfer}$: the network cannot fit any single target well, yielding biased distribution matching at every timestep. Compounding this, the student's intermediate predictions are imperfect and continuously evolving during training, the student thus optimizes against drifting, high-variance targets. Together these effects produce distribution mismatch, amplify error accumulation, and destabilize optimization.

These theoretical pathologies have concrete consequences. First, increasing the number of sampling steps often yields only marginal improvement, or even degradation~\cite{yin2024improved,DBLP:journals/corr/abs-2505-12674}, forcing existing methods to heavily rely on $\boldsymbol{\emph{real-image}}$ adversarial training to boost visual quality. Second, training is extremely costly: MMD runs each experiment on 256 TPU v5e chips~\cite{DBLP:journals/corr/abs-2406-04103}, while DMDv2 consumes 64 A100 GPUs per model~\cite{yin2024improved}. Third, the simultaneous objective lacks consistency across sampling budgets: a model trained for $N$-step inference produces high-quality output only at $N$ steps, while $1$ to $N{-}1$ steps degrade. Supporting $N$ budgets therefore requires training $N$ separate models, multiplying substantial training cost.

We introduce \textbf{Progressive Multistep Data-free Distillation (PMDD)}, a novel framework whose central mechanism directly removes the single-shared-network bottleneck. PMDD assigns a $\emph{dedicated}$ adapted network to each student score function $g_{\phi}\left(x_{t_{i}},t_{i}\right)$, so every stage is modeled accurately and free of negative transfer. To avoid the linear memory growth this would naively imply, PMDD adopts a $\emph{progressive}$ schedule that incrementally transforms a $K$-step student into a $K{+}1$-step student, building from $1$ up to $N$ steps. Each new stage is trained on samples drawn from the $\emph{previously distilled}$ student, which already approximates the relevant marginal well, replacing the drifting, high-variance targets of simultaneous training with stable ones, and sharply accelerating convergence. Finally, an $\emph{unforgetting loss}$ regularizes against the previous student, preserving earlier-step behavior so that one PMDD model produces consistent, high-quality output across $\emph{all}$ inference budgets, a property no prior data-free method offers.

We evaluate our approach across various tasks, including conditional image generation on CIFAR-10~\citep{krizhevsky2009learning}, ImageNet 64$\times$64~\citep{russakovsky2015imagenet}, and zero-shot text-to-image generation on MS COCO 512$\times$512~\citep{DBLP:conf/eccv/LinMBHPRDZ14}. PMDD improves with the number of steps and matches or outperforms data-free few-step baselines at every multistep budget, including FS-SiD~\citep{DBLP:journals/corr/abs-2505-12674}, reaching \textbf{FID 7.76} on MS-COCO 2014-30k, while supporting three inference budgets at 4--5$\times$ lower training compute than data-free DMDv2. We further show that a single PMDD model maintains strong consistency across inference budgets, with higher step counts progressively refining detail while preserving global structure, and that the framework generalizes to higher-resolution $1024\times1024$ setting.

%% file: sections/prelim.tex
\subsection{Data-free One-step Diffusion Distillation}
\label{subsec:Data-free-One-step-Diffusion}

Data-free one-step diffusion distillation aims to train a generator $g_{\phi}\left(x_{T},T\right)$ (the \emph{student}) that directly maps Gaussian noise $x_{T}\sim\Normal\left(0,\mathrm{I}\right)$ to data samples $x_{0}$ under the guidance of a pretrained diffusion model (the \emph{teacher}). A common strategy is to align the student distribution $p_{\phi}\left(x_{0}\right)$ with the teacher distribution $p_{\theta}\left(x_{0}\right)$. This is typically achieved by minimizing the \emph{reverse} KL divergence:
\begin{equation}
\Loss_{\text{KL}}\left(\phi\right):=D_{\text{KL}}\left(p_{\phi}\left(x_{0}\right)\|p_{\theta}\left(x_{0}\right)\right),\label{eq:match_x0_distribution}
\end{equation}
which differs from the forward KL divergence $D_{\text{KL}}\left(p_{\mathcal{D}}\left(x_{0}\right)\|p_{\phi}\left(x_{0}\right)\right)$ in data-dependent distillation.

In practice, this objective is optimized through its variational upper bound~\citep{ho2020denoising,song2020score}, which reduces to the denoised-sample matching (DM) loss:
\begin{equation}
\Loss\left(\phi\right):=\Expect_{x_{T},x_{0|\phi,T},t,\epsilon,x_{t}}\left[w_{x}\left(t\right)\left\Vert x_{\phi}\left(x_{t},t\right)-x_{\theta}\left(x_{t},t\right)\right\Vert ^{2}_{2}\right]\label{eq:denoised_matching}
\end{equation}
where $x_{T}\sim\mathcal{N}\left(0,\mathrm{I}\right)$, $x_{0|\phi,T}:=g_{\phi}\left(x_{T},T\right)$, $t\sim\mathcal{U}\left(1,T\right)$, $\epsilon\sim\mathcal{N}\left(0,\mathrm{I}\right)$, $x_{t}=a_{t}x_{0|\phi,T}+\sigma_{t}\epsilon$. Here, $w_{x}\left(t\right)>0$ denotes a time-dependent weighting factor under the denoised-sample parameterization.

A key difficulty in minimizing this loss is that $x_{\phi}\left(x_{t},t\right)$ is not directly accessible. A simple workaround is to adopt a surrogate loss~\citep{poole2022dreamfusion}:
\[
\tilde{\Loss}_{\text{DM}}\left(\phi\right):=\Expect_{x_{T},x_{0|\phi,T},t,\epsilon,x_{t}}\left[w_{x}\left(t\right)\left\Vert \bar{x}_{0|\phi,T}-x_{\theta}\left(x_{t},t\right)\right\Vert ^{2}_{2}\right]
\]
where $\bar{x}_{0|\phi,T}:=\text{sg}\left(g_{\phi}\left(x_{T},T\right)\right)$ and $\text{sg}\left(\cdot\right)$ denotes the stop-gradient operator.

However, this loss is inherently biased, resulting in samples with degraded quality and limited diversity, as observed in~\citep{wang2024prolificdreamer}. A better approach is to find a good approximation of $x_{\phi}\left(x_{t},t\right)$. This can be done by training an adapted denoising network $x_{\varphi}$ on samples generated by $g_{\phi}$ using the following objective:
\begin{equation}
\Loss^{\text{adapt}}_{\text{DM}}\left(\varphi\right):=\Expect_{x_{T},x_{0|\phi,T},t,\epsilon,x_{t}}\left[w_{x}\left(t\right)\left\Vert x_{\varphi}\left(x_{t},t\right)-x_{0|\phi,T}\right\Vert ^{2}_{2}\right]\label{eq:VSD_adapted_loss}
\end{equation}
After training $x_{\varphi}$, we update $g_{\phi}$ using a version of $\Loss_{\text{DM}}\left(\phi\right)$ with $x_{\phi}\left(x_{t},t\right)$ replaced by $x_{\varphi}\left(x_{t},t\right)$:
\begin{equation}
\Loss^{\text{stu}}_{\text{DM}}\left(\phi\right)\approx\Expect_{x_{T},x_{0|\phi,T},t,\epsilon,x_{t}}\left[w_{x}\left(t\right)\left\Vert x_{\varphi}\left(x_{t},t\right)-x_{\theta}\left(x_{t},t\right)\right\Vert ^{2}_{2}\right]\label{eq:VSD_student_loss}
\end{equation}
To stabilize training, prior works~\citep{wang2024prolificdreamer,nguyen2024swiftbrush,yin2024one} replace the full gradient $\nabla_{\phi}\Loss^{\text{stu}}_{\text{DM}}\left(\phi\right)$ with a modified gradient:
\begin{equation}
\tilde{\nabla}_{\phi}\Loss^{\text{stu}}_{\text{DM}}\left(\phi\right):=\Expect\left[w_{x}\left(t\right)\left(x_{\varphi}\left(x_{t},t\right)-x_{\theta}\left(x_{t},t\right)\right)\frac{\partial x_{0|\phi,T}}{\partial\phi}\right]\label{eq:VSD_student_grad}
\end{equation}
In practice, $x_{\varphi}$ and $g_{\phi}$ are optimized alternately via $\Loss^{\text{adapt}}_{\text{DM}}\left(\varphi\right)$ and $\tilde{\nabla}_{\phi}\Loss^{\text{stu}}_{\text{DM}}\left(\phi\right)$. Recently, Zhou et al.\ (2024)~\cite{zhou2024score} introduced the Score Identity Distillation (SiD) loss, which provides richer supervisory signals for training the student model:

\begin{multline}
\Loss^{\text{stu}}_{\text{SiD}}\left(\phi\right)=\Expect_{x_{0|\phi,T},t,\epsilon,x_{t}}\Big[w_{x}\left(t\right)\left(x_{\theta}\left(x_{t},t\right)-x_{\varphi}\left(x_{t},t\right)\right)^{\top}\\
\cdot\left(x_{\theta}\left(x_{t},t\right)-x_{0|\phi,T}\right)\Big]\label{eq:SiD_student_loss}
\end{multline}

\subsection{Data-free Multi-step Diffusion Distillation}
\label{subsec:Data-free-Multi-step-Diffusion}

A one-step student is often limited to model complex data distributions effectively. To overcome this limitation, multi-step diffusion distillation trains an $N$-step student model $g_{\phi}$ to generate data samples sequentially from $N$ time steps $t_{1}$, $t_{2}$, ..., $t_{N}$, where $0<t_{N}<t_{N-1}<\ldots<t_{1}=T$.

Previous multi-step approaches such as DMDv2~\citep{yin2024improved} and FS-SiD~\citep{DBLP:journals/corr/abs-2505-12674} learn the student mapping $g_{\phi}\left(x_{t_{i}},t_{i}\right)\mapsto x_{0}$ for all time steps $\left\{ t_{i}\right\} ^{N}_{i=1}$ simultaneously, following a strategy similar to one-step distillation. The noisy input $x_{t_{i}}$ is constructed by first sampling $x_{t_{1}}\sim\Normal\left(0,\mathrm{I}\right)$, and then recursively applying:
\begin{equation}
\bar{x}_{0|\phi,t_{k}}=\text{sg}\left(g_{\phi}\left(x_{t_{k}},t_{k}\right)\right),\ \ x_{t_{k+1}}=a_{t_{k+1}}\bar{x}_{0|\phi,t_{k}}+\sigma_{t_{k+1}}\epsilon\label{eq:multistep_noisy_sample}
\end{equation}
for $k=1,...,i-1$. Finally, $x_{0|\phi,t_{i}}:=g_{\phi}\left(x_{t_{i}},t_{i}\right)$ is computed without stopping gradients. The student objective in DF-DMDv2 resembles the DM loss in Eq.~\ref{eq:VSD_student_loss}:
\begin{equation}
\mathcal{L}^{\text{stu}}_{\text{DMDv2}}\left(\phi\right)=\Expect_{x_{0|\phi,t_{i}},i,x_{t}}\left[w_{x}\left(t\right)\left\Vert x_{\varphi}\left(x_{t},t\right)-x_{\theta}\left(x_{t},t\right)\right\Vert ^{2}_{2}\right]\label{eq:DF-DMDv2-loss}
\end{equation}
where $i\sim\mathcal{U}\left(1,N\right)$. FS-SiD considers an alternative objective based on the SiD loss as defined in Eq.~\ref{eq:SiD_student_loss}.

Despite simplicity in formulation, these objectives are prone to both bias and high variance. First, they rely on a single adapted network $x_{\varphi}$ to approximate multiple score functions corresponding to different data distributions produced by $g_{\phi}\left(x_{t_{i}},t_{i}\right)$ ($i=1,...,N$). Second, since the student's outputs at intermediate time steps $t_{k}$ ($k=1,...,i-1$) are imperfect and continually evolving during training, the resulting noisy samples $x_{t_{i-1}}$ deviate from the marginal distribution $p_{\theta}\left(x_{t_{i-1}}\right)$ and exhibit high variance. These factors collectively cause distribution mismatch and error accumulation, leading to slow convergence and unstable training.

%% file: sections/method.tex
\begin{figure*}[t]
\centering
\includegraphics[width=0.8\linewidth]{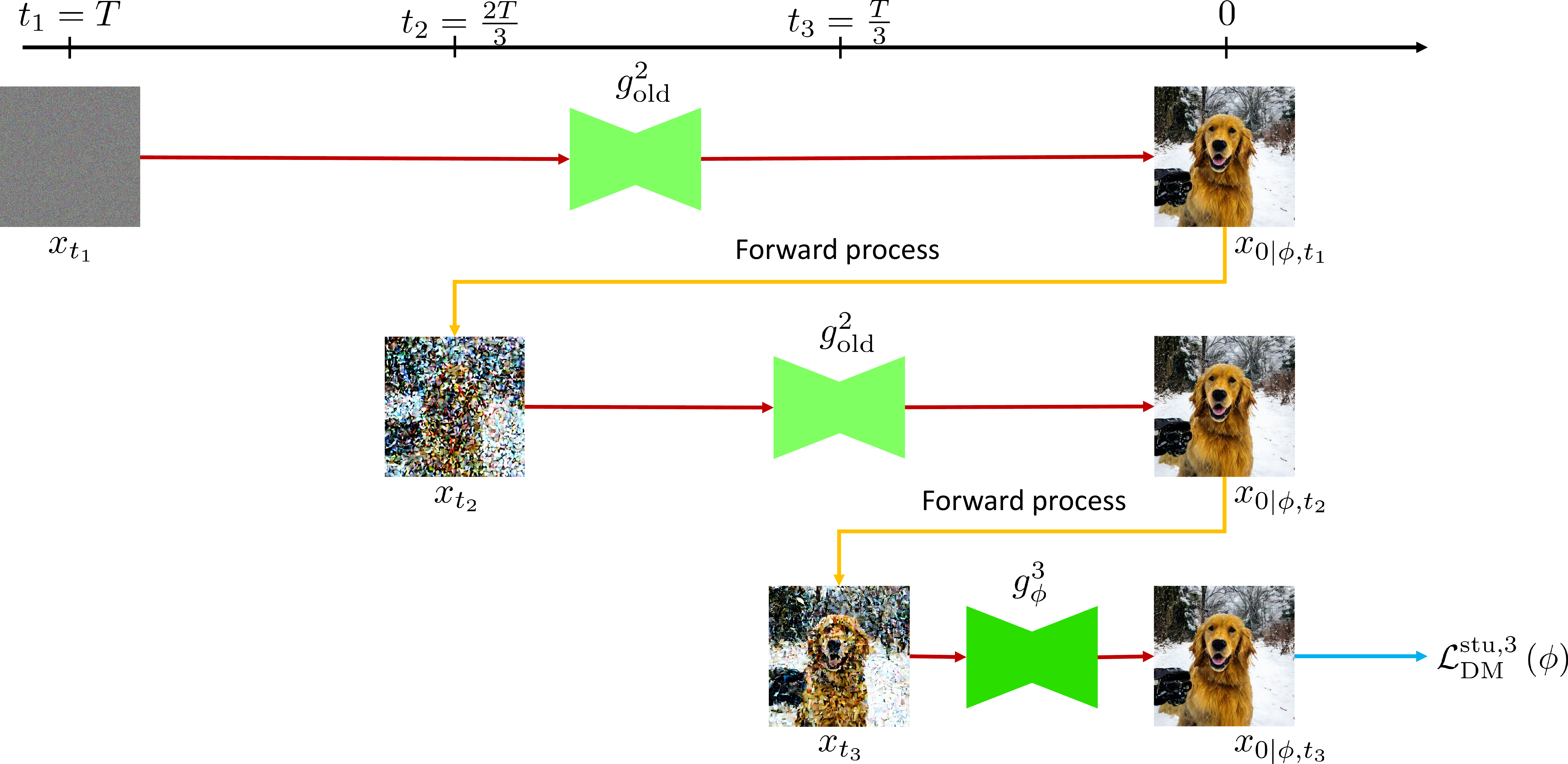}
\caption{Overview of our progressive distillation framework at stage 3. For clarity, the unforgetting loss and adapted network $x_{\varphi}$ are omitted from the figure.}
\end{figure*}

\subsection{Progressive Data-free Multi-step Distillation}

To overcome the limitations of existing multi-step distillation approaches discussed in Section~\ref{subsec:Data-free-Multi-step-Diffusion}, we propose a \emph{progressive multi-step diffusion distillation framework} instead of jointly learning the student mapping across all $N$ steps. In our approach, the student $g_{\phi}$ is distilled sequentially in multiple stages, progressing from $1$ to $N$.

Specifically, at the first stage ($t_{1}=T$), we train a one-step student $g^{1}_{\phi}$ using Gaussian noise as inputs, following the one-step distillation framework described in Section~\ref{subsec:Data-free-One-step-Diffusion}. Suppose that after stage $i-1$, we have obtained an $(i-1)$-step student $g^{i-1}_{\phi}$ capable of generating clean samples from step $t_{1}$ up to step $t_{i-1}$. At stage $i$, we leverage this previously distilled student to learn $g^{i}_{\phi}$, which extends generation capability up to step $t_{i}$. This procedure is repeated progressively until the final $N$-step student $g^{N}_{\phi}$ is learned.

To sample $x_{t_{i}}$, we first draw $x_{t_{1}}\sim\Normal\left(0,\mathrm{I}\right)$ and recursively apply
\begin{equation}
\bar{x}_{0|\phi,t_{k}}=\text{sg}\left(g^{i-1}_{\text{old}}\left(x_{t_{k}},t_{k}\right)\right),\ \ x_{t_{k+1}}=a_{t_{k+1}}\bar{x}_{0|\phi,t_{k}}+\sigma_{t_{k+1}}\epsilon,\label{eq:progressive_multistep_noisy_samples}
\end{equation}
for $k$ running from $1$ to $i-1$. Unlike Eq.~\ref{eq:multistep_noisy_sample}, we use the previously distilled student $g^{i-1}_{\text{old}}$ instead of the current $g^{i}_{\phi}$ to construct $x_{t_{i}}$. Since $g^{i-1}_{\text{old}}$ has already been well optimized, the $i$-step student $g^{i}_{\phi}\left(x_{t_{i}},t_{i}\right)$ is trained on a stable, low-variance distribution close to $p_{\theta}\left(x_{t_{i}}\right)$ rather than a drifting one, significantly improving training stability~\cite{DBLP:journals/nature/MnihKSRVBGRFOPB15}.

Let $x_{0|\phi,t_{i}}:=g^{i}_{\phi}\left(x_{t_{i}},t_{i}\right)$. At stage $i$, we alternately optimize the adapted network $x_{\varphi}$ and the student $g^{i}_{\phi}$ using the following losses:
\begin{equation}
\Loss^{\text{adapt},i}_{\text{DM}}\left(\varphi\right)=\Expect_{x_{0|\phi,t_{i}},t,x_{t}}\left[w_{x}\left(t\right)\left\Vert x_{\varphi}\left(x_{t},t\right)-x_{0|\phi,t_{i}}\right\Vert ^{2}_{2}\right]\label{eq:adapted_loss_ours}
\end{equation}
\begin{equation}
\Loss^{\text{stu},i}_{\text{DM}}\left(\phi\right)=\Expect_{x_{0|\phi,t_{i}},t,x_{t}}\left[w_{x}\left(t\right)\left\Vert x_{\varphi}\left(x_{t},t\right)-x_{\theta}\left(x_{t},t\right)\right\Vert ^{2}_{2}\right]\label{eq:student_loss_ours}
\end{equation}
The parameters of the student and adapted network at stage $i$ are initialized from those learned at stage $i-1$.

Compared with simultaneous distillation methods~\citep{yin2024improved,DBLP:journals/corr/abs-2511-22677,DBLP:journals/corr/abs-2505-12674}, our progressive approach assigns a dedicated adapted network at each stage $i$ to approximate the score function of distribution induced by $g^{i}_{\phi}\left(x_{t_{i}},t_{i}\right)$. This specialization is crucial in few-step regime: distant timesteps $t_{i}$ yield score functions that differ substantially across stages, and forcing such divergent targets into one parameter space $\Phi$ causes negative transfer~\cite{DBLP:conf/icml/StandleyZCGMS20}. This incurs a per-stage distribution bias $\Delta_{i}$ inflating the student gradient error in Eq.~\ref{eq:student_loss_ours} as formalized below.

\begin{proposition}[Per-stage decoupling lowers the aggregate score-matching error]
\label{prop:decoupling}
Let $\Loss_{i}\!\left(\varphi\right):=\Loss^{\text{adapt},i}_{\text{DM}}\!\left(\varphi\right)$ be the stage-$i$ score-matching loss (Eq.~\ref{eq:adapted_loss_ours}) with per-stage optimum $\varphi^{\star}_{i}\in\arg\min_{\varphi\in\Phi}\Loss_{i}\!\left(\varphi\right)$, and let $\varphi^{\star}\in\arg\min_{\varphi\in\Phi}\sum_{i}\Loss_{i}\!\left(\varphi\right)$ be the shared optimum of simultaneous distillation. Writing $\Delta_{i}:=\Loss_{i}\!\left(\varphi^{\star}\right)-\Loss_{i}\!\left(\varphi^{\star}_{i}\right)$,
\[
\underbrace{\sum^{N}_{i=1}\min_{\varphi_{i}\in\Phi}\Loss_{i}\!\left(\varphi_{i}\right)}_{\text{progressive}}\;\le\;\underbrace{\min_{\varphi\in\Phi}\sum^{N}_{i=1}\Loss_{i}\!\left(\varphi\right)}_{\text{simultaneous}},\,\,\text{gap}=\sum^{N}_{i=1}\Delta_{i}\;\ge\;0.
\]
\end{proposition}

The gap is $\emph{strict}$ whenever two stages have distinct marginals $p^{i}_{\phi}\ne p^{j}_{\phi}$, which holds generically in the few-step regime where score functions differ substantially across stages. Proposition proof is provided in Appendix A.1.

\subsection{Mitigating Catastrophic Forgetting}
\label{subsec:Mitigating-Catastrophic-Forgetti}

Solely optimizing the student objective in Eq.~\ref{eq:student_loss_ours} leads to severe catastrophic forgetting. This occurs because, as training progresses, the student increasingly concentrates on learning the $t_{i}\rightarrow0$ mapping, while neglecting previously learned mappings from earlier steps $t_{k}$ ($k<i$) -- which are excluded from gradient updates due to the stop-gradient operation. Consequently, using the $i$-step student to generate samples in fewer steps $k$ ($k\leq i$) noticeably degrades both semantic and visual fidelity.

To address this issue, we augment Eq.~\ref{eq:student_loss_ours} with an \emph{unforgetting regularizer}:
\[
\Loss^{\text{stu},i}_{\text{unf}}\left(\phi\right)=\sum^{i-1}_{k=1}\mathbb{E}_{x_{T},x_{t_{k}}}\left[\left\Vert g^{i}_{\phi}\left(x_{t_{k}},t_{k}\right)-g^{k}_{\text{old}}\left(x_{t_{k}},t_{k}\right)\right\Vert ^{2}_{2}\right]
\]
This enforces consistency between current student $g^{i}_{\phi}$ and the previously distilled students at their respective time steps, thereby preserving the learned mappings $t_{k}\rightarrow0$ for $k=1,...,i-1$. However, this formulation incurs a significant memory overhead that grows linearly with $i$, because it requires retaining all $i-1$ prior student models. To improve scalability, we adopt a more memory-efficient variant:
\begin{equation}
\Loss^{\text{stu},i}_{\text{unf}}\left(\phi\right)=\sum^{i-1}_{k=1}\mathbb{E}_{x_{T},x_{t_{k}}}\left[\left\Vert g^{i}_{\phi}\left(x_{t_{k}},t_{k}\right)-g^{i-1}_{\text{old}}\left(x_{t_{k}},t_{k}\right)\right\Vert ^{2}_{2}\right]\label{eq:eq:unforget_loss}
\end{equation}

This alternative leverages the most recently distilled student $g^{i-1}_{\text{old}}$ as the reference for all earlier steps, rather than storing all prior models. Since $g^{i-1}_{\text{old}}$ already maintains high-quality mappings from $t_{1}$, ..., $t_{i-2}$ to 0, a triangle-inequality bound (Appendix A.2) decomposes the deviation from every earlier student into at most $N-1$ per-stage residuals, reducing memory complexity to $\mathcal{O}(1)$ in the number of stages (Section~\ref{subsec:unforget_single_checkpoint}).

Overall, the student is optimized with the following loss:
\[
\Loss^{\text{stu},i}\left(\phi\right)=\Loss^{\text{stu},i}_{\text{DM}}\left(\phi\right)+\text{\ensuremath{\alpha}}\Loss^{\text{stu},i}_{\text{unf}}\left(\phi\right)
\]

%% file: sections/exp.tex
\begin{table*}[t]
\centering
\begin{minipage}{0.45\linewidth}
\centering
\resizebox{\textwidth}{!}{%
\begin{tabular}{lccc}
\toprule
Method & NFE $\left(\downarrow\right)$ & FID $\left(\downarrow\right)$ & CLIP $\left(\uparrow\right)$ \\
\midrule
\multicolumn{4}{c}{\textbf{Teacher Model}} \\
SD v1.5 (CFG=3) \citep{rombach2022high} & 50 & 8.78 & - \\
SD v1.5 (CFG=8) \citep{rombach2022high} & 50 & 13.45 & 0.322 \\
\midrule
\multicolumn{4}{c}{\textbf{One-step Distill.}} \\
DMD (CFG=3) \citep{yin2024one} & 1 & 11.49 & - \\
InstaFlow-1.7B \citep{liu2023instaflow} & 1 & 11.8 & 0.309 \\
Rect. Diffusion \citep{DBLP:conf/iclr/Wang0HW025} & 1 & 10.21 & 0.310 \\
MMD \cite{DBLP:journals/corr/abs-2406-04103} & 1 & 9.63 & 0.321 \\
\textcolor{brown}{SwiftBrush \citep{nguyen2024swiftbrush}} & \textcolor{brown}{1} & \textcolor{brown}{16.67} & \textcolor{brown}{0.290} \\
\textcolor{brown}{Guided SiD \citep{DBLP:conf/iclr/ZhouWZH25}} & \textcolor{brown}{1} & \textcolor{brown}{8.15} & \textcolor{brown}{0.304} \\
\midrule
\multicolumn{4}{c}{\textbf{Multi-step Distill. (High CFG)}} \\
\multirow{2}{*}{\textcolor{brown}{FS-SiD (LSG=4.5) \citep{DBLP:journals/corr/abs-2505-12674}}} & \textcolor{brown}{1} & \textcolor{brown}{16.59} & \underline{\textcolor{brown}{0.317}} \\
 & \textcolor{brown}{4} & \textcolor{brown}{16.60} & \textbf{\textcolor{brown}{0.319}} \\
\multirow{2}{*}{\textcolor{brown}{PMDD (CFG=4.5) (ours)}} & \textcolor{brown}{1} & \underline{\textcolor{brown}{15.57}} & \underline{\textcolor{brown}{0.317}} \\
 & \textcolor{brown}{2} & \textbf{\textcolor{brown}{14.83}} & \textbf{\textcolor{brown}{0.319}} \\
\midrule
\multicolumn{4}{c}{\textbf{Multi-step Distill. (Low CFG)}} \\
\multirow{2}{*}{FS-SiD+GAN (Zero-CFG) \citep{DBLP:journals/corr/abs-2505-12674}} & 2 & 8.75 & 0.315 \\
 & 4 & 8.52 & 0.308 \\
\textcolor{brown}{DF DMDv2 (CFG=1.75) \citep{yin2024improved}} & \textcolor{brown}{1} & \textcolor{brown}{9.35} & \textbf{\textcolor{brown}{0.304}} \\
 & \textcolor{brown}{2} & \textcolor{brown}{10.44} & \textcolor{brown}{0.301} \\
 & \textcolor{brown}{3} & \textcolor{brown}{9.18} & \underline{\textcolor{brown}{0.303}} \\
\textcolor{brown}{PMDD (CFG=1.75) (ours)} & \textcolor{brown}{1} & \textcolor{brown}{10.41} & \textcolor{brown}{0.302} \\
 & \textcolor{brown}{2} & \textcolor{brown}{8.63} & \textcolor{brown}{0.300} \\
 & \textcolor{brown}{3} & \underline{\textcolor{brown}{8.50}} & \textcolor{brown}{0.302} \\
\textcolor{brown}{DF DMDv2 (CFG=1.0) \citep{yin2024improved}} & \textcolor{brown}{3} & \textcolor{brown}{9.02} & \textcolor{brown}{0.300} \\
\textcolor{brown}{PMDD (CFG=1.0) (ours)} & \textcolor{brown}{3} & \textbf{\textcolor{brown}{7.76}} & \textcolor{brown}{0.299} \\
\bottomrule
\end{tabular}
}
\label{tab:text2img_results}
\end{minipage}
\hspace{0.10\linewidth}
\begin{minipage}{0.30\linewidth}
\centering
\resizebox{\textwidth}{!}{%
\begin{tabular}{lcc}
\toprule
Method & NFE $\left(\downarrow\right)$ & FID $\left(\downarrow\right)$ \\
\midrule
\multicolumn{3}{c}{\textbf{Teacher Model}} \\
EDM \citep{karras2022elucidating} & 512 & 2.32 \\
\midrule
\multicolumn{3}{c}{\textbf{One-step Distillation}} \\
DFNO \citep{DBLP:conf/icml/ZhengNVAA23} & 1 & 7.83 \\
TRACT \citep{berthelot2023tract} & 1 & 7.43 \\
DMD \citep{yin2024one} & 1 & 2.62 \\
EMD-1 \citep{xie2024distillation} & 1 & 3.10 \\
EMD-16 \citep{xie2024distillation} & 1 & 2.60 \\
DMDv2 \citep{yin2024improved} & 1 & 1.28 \\
\textcolor{brown}{BOOT \citep{gu2023boot}} & \textcolor{brown}{1} & \textcolor{brown}{16.3} \\
\textcolor{brown}{SwiftBrush \citep{nguyen2024swiftbrush}} & \textcolor{brown}{1} & \textcolor{brown}{5.85} \\
\textcolor{brown}{Diff-Instruct \citep{DBLP:journals/corr/abs-2410-20898}} & \textcolor{brown}{1} & \textcolor{brown}{5.57} \\
\midrule
\multicolumn{3}{c}{\textbf{Multi-step Distillation}} \\
\multirow{2}{*}{PD \citep{salimans2022progressive}} & 1 & 15.39 \\
 & 2 & 8.95 \\
\multirow{2}{*}{CD \citep{song2023consistency}} & 1 & 6.20 \\
 & 2 & 4.70 \\
\multirow{2}{*}{MM Alt. \citep{DBLP:journals/corr/abs-2406-04103}} & 1 & 3.00 \\
 & 2 & 3.86 \\
\multirow{2}{*}{CTM \citep{DBLP:conf/iclr/KimLLMTUHME24}} & 1 & 1.92 \\
 & 2 & 1.73 \\
\textcolor{brown}{DF DMDv2 \citep{yin2024improved}} & \textcolor{brown}{1} & \textcolor{brown}{2.61} \\
 & \textcolor{brown}{2} & \textcolor{brown}{2.49} \\
\multirow{2}{*}{\textcolor{brown}{PMDD (ours)}} & \textcolor{brown}{1} & \underline{\textcolor{brown}{2.58}} \\
 & \textcolor{brown}{2} & \textbf{\textcolor{brown}{1.99}} \\
\bottomrule
\end{tabular}
}
\label{tab:imagenet_results}
\end{minipage}
\caption{\textbf{(a)}: Text-to-image generation performance on 30K COCO 2014 samples for Stable Diffusion v1.5 distillation. Data-free (DF) methods are marked in \textcolor{brown}{brown} with the best and second-best approaches highlighted in \textbf{bold} and \underline{underline}, respectively. \textbf{(b)}: Image generation results on ImageNet64; the teacher uses 512 NFE, as reported in~\cite{yin2024one}. }
\label{fig:main_table}
\end{table*}

We evaluate PMDD on two regimes. For $\emph{class-conditional}$ generation we use CIFAR-10 \citep{krizhevsky2009learning} and ImageNet64 \citep{russakovsky2015imagenet}, using pretrained teacher models from \citep{karras2022elucidating}. For $\emph{zero-shot text-to-image}$ generation, we distill Stable Diffusion v1.5 \citep{rombach2022high} for all methods and evaluate on 30K MS-COCO 2014 validation prompts \cite{DBLP:conf/eccv/LinMBHPRDZ14}. Following prior work settings \cite{yin2024improved,DBLP:conf/iclr/ZhouWZH25,DBLP:journals/corr/abs-2505-12674}, we report FID \citep{heusel2017gans}, CLIP \cite{radford2021learning}, and Human Preference Score v2 (HPSv2) \cite{DBLP:journals/corr/abs-2306-09341}. We use the same teacher (SDv1.5) and evaluation protocol for all methods; where a baseline is run at the same guidance scale, we report PMDD at that scale. Data-dependent and differently-resourced methods are included for completeness under their reported settings. All experiments use at most four GPUs. Implementation details are in Appendix B.

\subsection{Performance Comparison}
\label{subsec:text-to-image-generation}

PMDD is the strongest multistep data-free distiller in both guidance regimes (Table$~\ref{fig:main_table}a$). Under low guidance, three steps reach FID $7.76$, below the $50$-step teacher at CFG$\,=3$ (8.78), the data-free DMDv2 at the same CFG (9.02), and the data-dependent FS-SiD+GAN at $4$ steps, without real data or a GAN loss. Under high guidance, two PMDD steps match the best CLIP score of $\emph{four-step}$ FS-SiD (0.319) at a lower FID (14.83 vs.\ 16.60). On ImageNet64 (Table$~\ref{fig:main_table}$b), two-step PMDD surpasses its 512-NFE teacher (1.99 vs.\ 2.32) and data-free DMDv2.

\begin{figure}[h]
\centering
\includegraphics[width=\columnwidth]{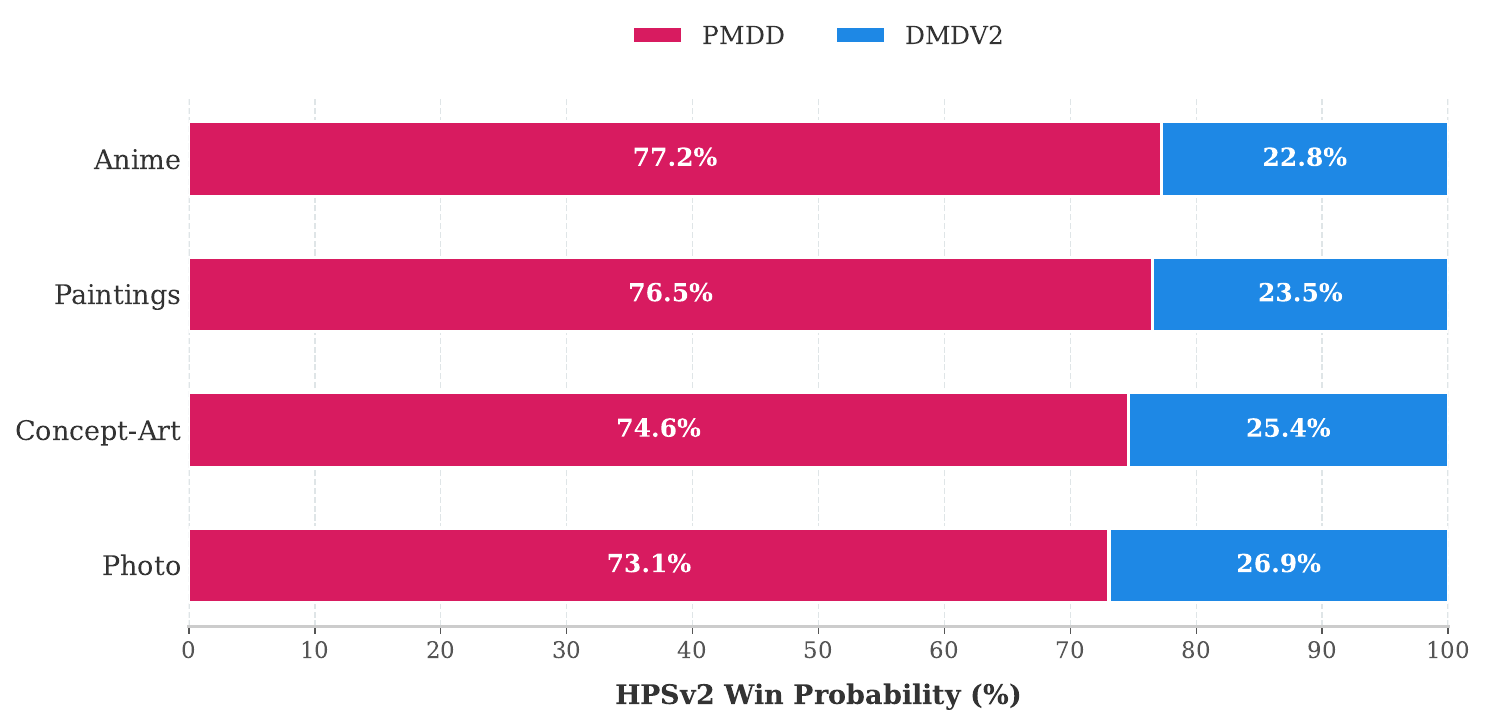}
\caption{HPSv2 win rate: PMDD vs. DF-DMDv2. }
\label{fig:hpsv2}
\end{figure}

These gains extend to perceptual quality. PMDD maintains a consistent $\approx$75\% HPSv2 win rate over DMDv2 across diverse visual themes (Figure$~\ref{fig:hpsv2}$), and produces more photo-realistic, better prompt-aligned images (Figure $~\ref{fig:consistency-image}$). DMDv2 hallucinates implausible scenes (water unnaturally surrounding the lighthouse, first row) while the teacher omits key components (``coastline'' in first row; ``waterfall'' in second row). PMDD avoids both failures, preserving both primary attributes and subtle context. This quality holds across step budgets with a single model, which we quantify in Section $\text{\ref{subsec:flexible_inference}}$

These results support our two core design choices: per-stage networks remove the excess score-matching error $\Delta_{i}$ of a shared network (Proposition$~\ref{prop:decoupling}$), and the frozen previous-step $g^{i-1}_{\text{old}}$ provides low-variance targets that stabilize optimization (Section$~\ref{subsec:external_sampling_ablation}$).

\subsection{Flexible Inference from a single model}
\label{subsec:flexible_inference}

A practical limitation of simultaneous distillation is that each target step count requires a separately trained model. PMDD removes this constraint: a single model serves any step budget, refining rather than recomposing.

\begin{figure*}[t]
\centering
\begin{minipage}{0.68\linewidth}
\centering
\includegraphics[width=\linewidth]{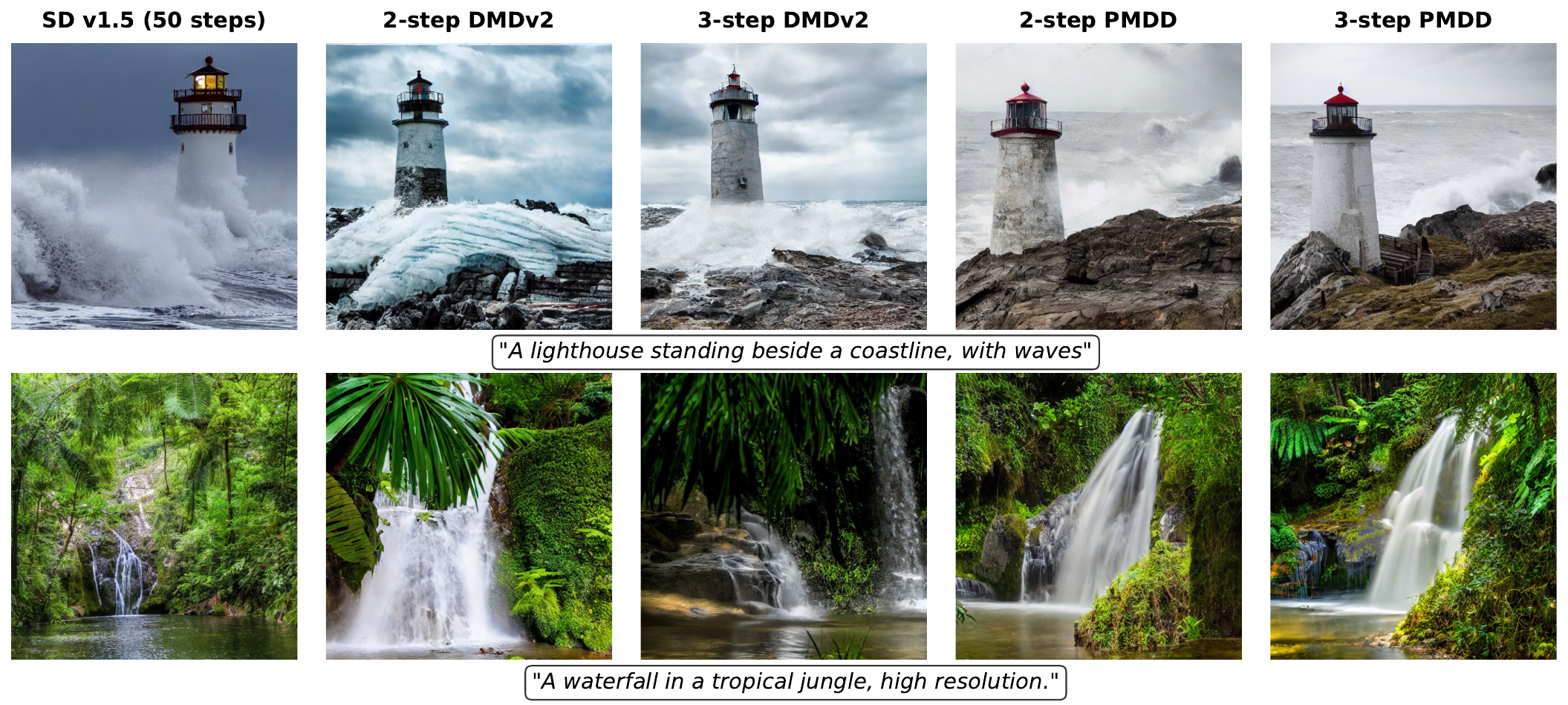}
\captionof{figure}{Text-to-image generation: SD v1.5 vs. few-step distillation methods}
\label{fig:consistency-image}
\end{minipage}\hspace{0.04\linewidth}\begin{minipage}{0.26\linewidth}
\centering
\resizebox{\textwidth}{!}{%
\begin{tabular}{lccc}
\toprule
Method & NFE=3 & NFE=2 & NFE=1 \\
\midrule
\textbf{2-step} & & & \\
GD \cite{meng2023distillation} & - & 33.25 & 108.21 \\
LCM \cite{luo2023latent} & - & 13.31 & 35.36 \\
SCFD \cite{DBLP:journals/corr/abs-2412-16906} & - & 11.46 & 11.91 \\
\textcolor{brown}{DMDv2} \cite{yin2024improved} & - & 10.44 & 17.46 \\
\textcolor{brown}{PMDD (ours)} & - & \textbf{8.63} & \textbf{11.67} \\
\midrule
\textbf{3-step} & & & \\
\textcolor{brown}{DMDv2} \cite{yin2024improved} & 9.18 & 13.71 & 26.49 \\
\textcolor{brown}{PMDD (ours)} & \textbf{8.50} & \textbf{10.07} & \textbf{12.65} \\
\bottomrule
\end{tabular}
}
\captionof{table}{FID across budgets. Data-free methods in \textcolor{brown}{brown}.}
\label{tab:tab:unified_model_results-1}
\end{minipage}
\end{figure*}

To quantify consistency, we fix the initial noise $x_{T}$ and measure how much a model's output changes as the step count varies. Table$~\ref{tab:training_efficiency-1}$ reports LPIPS and SSIM between step budgets where PMDD stays more consistent than DMDv2 on every step pair. Figures$~\ref{fig:teaser}$ and$~\ref{fig:consistency-image}$ show this qualitatively: DMDv2's layout shifts and degrades at low budgets, while PMDD holds its composition and sharpens detail. This stability is the direct effect of our unforgetting loss $\Loss^{\text{stu},i}_{\text{unf}}(\phi)$, which preserves global structure across budgets so that extra steps refine an image rather than recompose it.

\begin{table}[h]
\centering
\resizebox{0.65\columnwidth}{!}{%
\begin{tabular}{cccc}
\toprule
Methods & Step & LPIPS ($\downarrow$) & SSIM ($\uparrow$) \\
\midrule
DMDv2 & 1 vs 2 & 0.510 & 0.303 \\
 & 1 vs 3 & 0.534 & 0.309 \\
 & 2 vs 3 & 0.476 & 0.411 \\
\midrule
PMDD & 1 vs 2 & 0.352 & 0.651 \\
 & 1 vs 3 & 0.387 & 0.613 \\
 & 2 vs 3 & 0.360 & 0.637 \\
\bottomrule
\end{tabular}
}
\caption{Quantitative evaluation of cross-step structural consistency.}
\label{tab:training_efficiency-1}
\end{table}

Sampled with fewer steps, the 3-step DMDv2 collapses to FID 26.49 at NFE$\,=1$ (Table$~\ref{tab:tab:unified_model_results-1}$), whereas the 3-step PMDD degrades marginally, and at NFE$\,=2$ still beats DMDv2's dedicated 2-step model (10.07 vs.\ 10.44). One PMDD model thus covers the full 1--3 step range (further results in Appendix G).

\subsection{Training Efficiency}
\label{subsec:Training-Efficiency}

At inference, PMDD deploys a single student and discards all per-stage adapted networks and unforgetting regularizer. The savings are at training time: to support $N$ inference budgets, simultaneous methods train $N$ times, paying $N\times$ the cost, whereas PMDD pays once in the first stage and finetunes each later stage from the previous one (Table$~\ref{tab:training_efficiency}$). Supporting three budgets on SD v1.5 thus takes $4.3\times$ fewer synthetic samples and $4.4\times$ less time than DMDv2, with each additional step costing under 3\% of the first stage's samples (convergence in Appendix C). These gains stem from our progressive design, which reuses each stage to build the next for delivering multi-budget flexibility and high training efficiency.

\begin{table}[h]
\centering
\resizebox{0.90\columnwidth}{!}{%
\begin{tabular}{cccc}
\toprule
Step & \begin{tabular}[c]{@{}c@{}}DMDv2\\(Data / Time)\end{tabular} & \begin{tabular}[c]{@{}c@{}}PMDD\\(Data / Time)\end{tabular} & \begin{tabular}[c]{@{}c@{}}Efficiency\\(Data / Time)\end{tabular} \\
\midrule
1-step & 82M / 110 hrs & \textbf{56M / 65 hrs} & 1.5$\times$ / 1.7$\times$ \\
\midrule
2-step & 82M / 120 hrs & \textbf{0.32M / 12 hrs} & 256$\times$ / 10$\times$ \\
\midrule
3-step & 82M / 145 hrs & \textbf{1.3M / 8hrs} & $63\times$ / 18 $\times$ \\
\midrule
Total & 246M / 375 hrs & \textbf{57.6M / 85hrs} & $4.3\times$ / 4.4 $\times$ \\
\bottomrule
\end{tabular}
}
\caption{Training cost on SDv1.5 (million synthetic samples / wall-clock hours). DMDv2 trains one model per budget; PMDD's 2- and 3-step rows are incremental stage costs.}
\label{tab:training_efficiency}
\end{table}

\subsection{Ablation Studies}
\label{subsec:Ablation-Studies}

\subsubsection{Generalization to high-resolution}

To test scalability beyond U-Net $512\times512$, we apply PMDD to PixArt-$\alpha$, a Diffusion Transformer, at $1024\times1024$ and evaluate on MJHQ-30k following \cite{DBLP:conf/iccv/ChenXZYPCCHX25} (Table$~\ref{tab:Eval-on-MJHQ-30k}$). DF-DMDv2 again degrades at two steps, while PMDD improves with every step. With three steps, PMDD reaches the best FID and GenEval among the distilled models, ahead of TDM and Decoupled DMD at four steps, and surpasses its 20-step teacher in both (8.65 vs.\ 8.85; 0.54 vs.\ 0.51). Extending to larger backbones such as SDXL is left for future work.

\begin{table}[t]
\centering
\resizebox{0.90\columnwidth}{!}{%
\begin{tabular}{ccccc}
\toprule
Methods & Steps & FID ($\downarrow$) & CLIP ($\uparrow$) & GenEval (↑) \\
\midrule
PixArt$\alpha$ \cite{DBLP:conf/iclr/ChenYGYXWK0LL24} & 20 & 8.85 & 28.07 & 0.51 \\
SD3-medium \cite{DBLP:conf/icml/EsserKBEMSLLSBP24} & 28 & 11.92 & 27.83 & 0.62 \\
FLUX-dev \cite{DBLP:journals/corr/abs-2506-15742} & 50 & 10.15 & 27.47 & 0.67 \\
\midrule
DF-DMDv2 \cite{yin2024improved} & 1 & 10.94 & 27.62 & 0.45 \\
 & 2 & 12.08 & 27.68 & 0.44 \\
 & 3 & 10.83 & 27.72 & 0.49 \\
TDM \cite{DBLP:conf/iccv/LuoHSCT25} & 2 & 9.37 & 27.74 & 0.48 \\
 & 4 & 8.81 & \textbf{27.91} & 0.51 \\
DDMD \cite{DBLP:journals/corr/abs-2511-22677} & 2 & 9.01 & 27.82 & 0.50 \\
 & 4 & \underline{8.78} & 27.87 & \underline{0.52} \\
\midrule
\textbf{PMDD} & 1 & 10.06 & 27.83 & 0.48 \\
 & 2 & 9.11 & 27.88 & 0.50 \\
 & 3 & \textbf{8.65} & \underline{27.90} & \textbf{0.54} \\
\bottomrule
\end{tabular}
}
\caption{Data-free distillation of PixArt-$\alpha$ on $1024\times1024$ MJHQ-30k. Baselines are reproduced with official code and the same teacher; SD3-medium and FLUX-dev are reference generators. Best/second-best distilled results in \textbf{bold}/\underline{underline}.}
\label{tab:Eval-on-MJHQ-30k}
\end{table}

\subsubsection{Effect of the Unforgetting Loss Weight $\alpha$}
\label{subsec:unforget_weight_ablation}

We vary the unforgetting weight $\alpha$ in Figure $\text{\ref{fig:Effect-of-unforget_text_2_img}}$, which trades off preservation of earlier steps against current-step convergence. While high $\alpha$ values prioritize early-step preservation and offer greater training stability (Figure $\text{\ref{fig:Effect-of-unforget_text_2_img}}$b,c) , it results in slow convergence at the current step (Figure $\text{\ref{fig:Effect-of-unforget_text_2_img}}$a). Conversely, smaller weights prioritize the current step, at the cost of a faster rise in one- and two-step FID.

\begin{figure}[t]
\centering
\includegraphics[width=0.83\columnwidth]{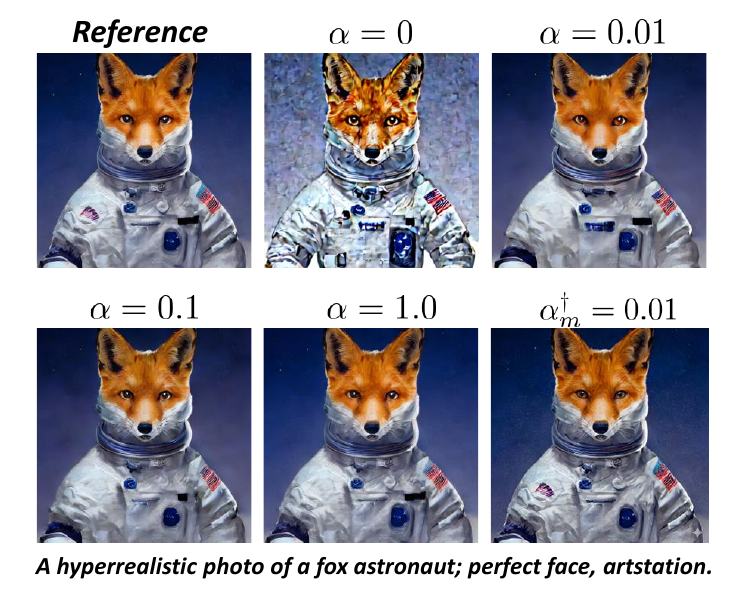}
\caption{Effect of $\alpha$ on 1-step sampling (NFE = 1) for a 3-step PMDD model. $\alpha^{\dagger}_{\text{m}}$ retains all previous checkpoints, whereas $\alpha$ retains only the latest. }
\label{fig:qualitative_of_unforget}
\end{figure}

Qualitatively, Figure $\text{\ref{fig:qualitative_of_unforget}}$ demonstrates that PMDD is highly robust across a wide range of values. While setting $\alpha=0$ results in a significant deterioration and over-saturation, there is no significant visual difference for any $\alpha\geq0.1$, all successfully preserving overall fidelity and structure relative to the reference image.

\subsubsection{Effect of Using the Most Recent Distilled Student}
\label{subsec:unforget_single_checkpoint}

\begin{figure*}[t]
\centering
\includegraphics[width=0.8\linewidth]{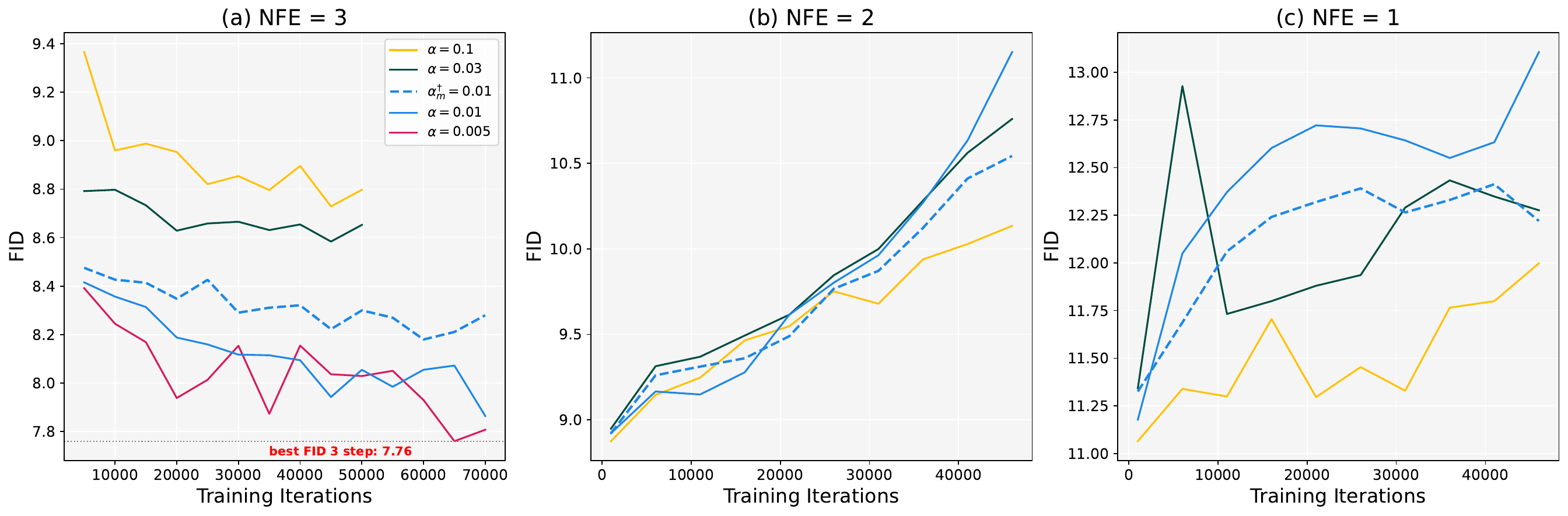}
\caption{Effect of unforget weight $\alpha$ on 3-step inference for COCO 2014. $\alpha^{\dagger}_{m}$ preserves all previous checkpoints, while $\alpha$ preserves only the latest.}
\label{fig:Effect-of-unforget_text_2_img}
\end{figure*}

We compare the effectiveness of regularizing against the last distilled checkpoint versus all previous checkpoints in Figure $\text{\text{\ref{fig:qualitative_of_unforget}}}$. Both strategies produce nearly identical reconstructions, closely matching the reference image. Quantitatively (Figure $\text{\ref{fig:Effect-of-unforget_text_2_img}}$), the latest-only reference reaches lower 3-step FID, whereas the full history retains lower 1- and 2-step FID. These results support our design choice: regularizing against multiple checkpoints over-constrains the student, while using the last checkpoint provides greater flexibility to better match the current step distribution.

\subsubsection{Effect of Previous-Step Sampling}
\label{subsec:external_sampling_ablation}

\begin{figure}[t]
\centering
\includegraphics[width=\columnwidth]{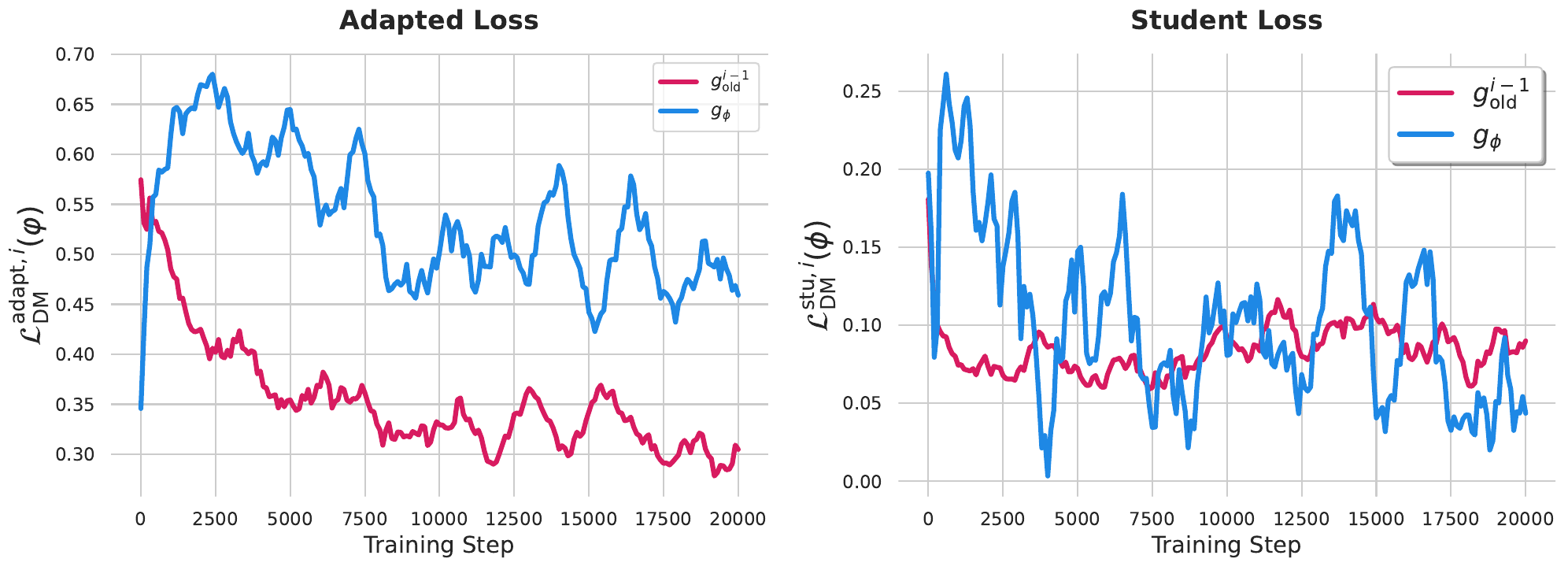}
\caption{Ablation Study of using the previously distilled model $g^{i-1}_{old}$ for sampling $x_{t_{i}}$ on ImageNet64.}
\label{fig:previous_step_sampling}
\end{figure}

We ablate the sampling source for $x_{t_{i}}$ in Figure $\text{\ref{fig:previous_step_sampling}}$. Sampling from the previously-distilled model $g^{i-1}_{\text{old}}$ yields significant and consistent FID gains on ImageNet64 over using the evolving student model $g_{\phi}$. This produces a noticeably smoother, lower-variance adapted and student loss throughout training. This result confirms that sampling from $g^{i-1}_{\text{old}}$ effectively reduces the high variance inherent in evolving student outputs, thereby mitigating distribution shift and stabilizing optimization.

%% file: sections/relate.tex
One-step distillation achieves great performance on small- and medium-sized resolution~\cite{nguyen2024swiftbrush,DBLP:conf/iclr/ZhouWZH25,wang2022diffusion,zhou2024score,yin2024one,DBLP:conf/cvpr/0003ZXH24} building on Score Distillation Sampling~\cite{DBLP:conf/iclr/PooleJBM23} and Variational Score Distillation~\cite{wang2024prolificdreamer}. Several methods have been designed to extend distillation from one-step to multistep settings for higher resolution. However, they naively extend from the one-step distillation approach, which uses a single adapted network to learn the score function corresponding to data distributions induced by $g_{\phi}\left(x_{t_{i}},t_{i}\right)$ ($i=1,...,N$). Because these student distributions are themselves imperfect and evolve during training, the shared network must jointly model multiple unstable targets, causing distribution mismatch, error accumulation, and unstable optimization (Section~\ref{subsec:Data-free-Multi-step-Diffusion}). Therefore, quality improves only marginally with additional sampling steps and, in some cases, fails to improve. In contrast, our key novelty is to assign a distinct adapted network to each student to each stage through our progressive training mechanism. Our extensive experiment validates this design, demonstrating more training-efficient and superior performance.

A separate line of work distills by $\emph{matching the teacher's sampling trajectory}$, which regresses the student onto points along the teacher's deterministic ODE trajectory. Typically, they require real data/images or teacher's trajectories for learning~\cite{salimans2022progressive,luhman2021knowledge,liu2022flow,liu2022rectified,DBLP:conf/iclr/KimLLMTUHME24,luo2023latent,song2023consistency,gu2023boot}. PMDD differs fundamentally: it is data-free and distribution-matching, aligning student and teacher distributions through a learned score estimate without ever referencing a real sample or a teacher trajectory. This distinction is sharpest against Progressive Distillation (PD)~\cite{salimans2022progressive}, with which our method shares a ``progressive design''. PD's progression $\emph{halves}$ the number of sampling steps, compressing a $2k$-step teacher into a $k$-step student, so each student is upper-bounded by its teacher and trades quality for speed. PMDD instead distills a teacher into a one-step student, then incrementally extends a $k$-step student into $k+1$ step one, improving quality as steps increase.

%% file: sections/discuss.tex
We introduced Progressive Multistep Data-free Distillation (PMDD). Simultaneous distillation fits a single adapted network to all steps at once, on inputs produced by a student that is itself changing. PMDD instead matches each stage with its own adapted network on inputs from the frozen previous student, and an unforgetting loss keeps a single model usable at every budget. Across CIFAR-10, ImageNet $64\times64$, COCO $512\times512$ and MJHQ-30K $1024\times1024$, PMDD matches or outperforms data-free baselines at every multistep budget and improves with every added step on both U-Net and Transformer backbones, at 4--5$\times$ lower training cost than one data-free DMDv2 model per budget. We focus on the few-step regime ($N<4$); beyond it, sampling approaches the teacher's inference cost, diminishing the advantage of distillation.

%% file: sections/appendix.tex
\section{Proofs}

\subsection{Proof of Proposition~\ref{prop:decoupling}}
\label{app:proof-decoupling}

Let $\Loss_{i}(\varphi):=\Loss^{\text{adapt},i}_{\text{DM}}(\varphi)$ be the stage-$i$ score-matching loss (Eq.~\ref{eq:adapted_loss_ours}), with shared optimum $\varphi^{\star}\in\arg\min_{\varphi}\sum_{i}\Loss_{i}(\varphi)$, per-stage optima $\varphi^{\star}_{i}\in\arg\min_{\varphi}\Loss_{i}(\varphi)$, and $\Delta_{i}:=\Loss_{i}(\varphi^{\star})-\Loss_{i}(\varphi^{\star}_{i})$. Since $\varphi^{\star}_{i}$ minimizes $\Loss_{i}$, the aggregate gap is immediate:
\[
\min_{\varphi}\sum_{i}\Loss_{i}(\varphi)-\sum_{i}\min_{\varphi}\Loss_{i}(\varphi)=\sum_{i}\Delta_{i}\ \ge0.
\]

The shared optimum is strictly worse than the per-stage optima ($\Delta_{i}>0$) exactly when the stages impose conflicting targets on $\varphi$. This holds in few-step regime where the score functions of the student distribution induced by $g^{i}_{\phi}\left(x_{t_{i}},t_{i}\right)$ at distant timesteps $t_{i}$ differ substantially. Therefore, no single $\varphi\in\Phi$ fits all of them at once. Forcing one network onto such heterogeneous targets is the classic source of $\emph{negative transfer}$ in multi-task learning, where the shared solution compromises across tasks \cite{DBLP:conf/icml/StandleyZCGMS20}. Hence $\sum_{i}\Delta_{i}>0$.

The student update is driven by the residual $x_{\varphi}-x_{\theta}$, so its correctness hinges on which $x_{\varphi}$ is used. The correct gradient at stage $i$ uses the stage-$i$ optimum $x_{\varphi^{\star}_{i}}$; simultaneous training instead uses the shared $x_{\varphi^{\star}}$, so the realized update is biased by $x_{\varphi^{\star}}-x_{\varphi^{\star}_{i}}$. This bias is nonzero precisely when $\Delta_{i}>0$, i.e. when the shared network is suboptimal at stage $i$. This persists as a systematic error in the reverse-KL distribution match whereas per-stage decoupling uses $x_{\varphi^{\star}_{i}}$ directly, eliminating it at every stage and tightening distribution matching across all stages.

\subsection{Single-reference proxy bounds all earlier-step deviations}
\label{app:proof-proxy}

The single-reference loss (Eq.~\ref{eq:eq:unforget_loss}) penalizes the deviation of the current student $g^{i}_{\phi}$ from the most recent student $g^{i-1}_{\text{old}}$, rather than from all prior students. We show this suffices to control the deviation from every earlier student. The $L_{2}$ triangle inequality shows this still controls the deviation from every earlier student: for any $k<i$, at $(x_{t_{k}},t_{k})$,
\[
\bigl\| g^{i}_{\phi}-g^{k}_{\text{old}}\bigr\|_{2}\;\le\underbrace{\bigl\| g^{i}_{\phi}-g^{i-1}_{\text{old}}\bigr\|_{2}}_{\text{minimized by Eq.~\ref{eq:eq:unforget_loss}}}+\underbrace{\bigl\| g^{i-1}_{\text{old}}-g^{k}_{\text{old}}\bigr\|_{2}}_{\text{controlled at stage }i-1},
\]

The first term is driven small by the current stage's loss, and the second is the deviation already constrained when $g^{i-1}_{\text{old}}$ was trained in previous stage. Hence matching the latest student alone controls the deviation from all earlier ones, with $\mathcal{O}(1)$ memory in the number of stages. Iterating this bound, the deviation from any prior student decomposes into at most $N-1$ per-stage residuals, which do not vanish: in practice, earlier budgets degrade gradually but do not collapse (Fig.~\ref{fig:Effect-of-unforget_text_2_img} and Table~\ref{tab:tab:unified_model_results-1}). We target the few-step regime $N<4$, following prior distillation work \cite{yin2024improved,DBLP:journals/corr/abs-2511-22677,DBLP:journals/corr/abs-2505-12674}, because increasing the sampling steps beyond it approaches the teacher's inference steps (around 20), diminishing the advantage of distillation. There, at most two residuals accumulate, which is sufficient in practice.

\section{Implementation Details}
\label{subsec:Implementation-Details-1}

\subsection{CIFAR-10}

We distill a one-step student model from the EDM Pretrained model \cite{karras2022elucidating}, using edm-cifar10-32x32-cond-vp for class-conditional image generation. Training is performed with AdamW optimizer (learning rate of 1e-5, a weight decay 0.01, and beta parameters $\left(0.9,0.999\right)$), a total batch size of 256 on 4 A100 GPUs for 200K iterations. We use the baseline DMD as our codebase and adopt a two-time update rule: the adapted network is updated 10 times per student update, and the weighting factor $w\left(t\right)$ from \cite{yin2024one} is applied to stabilize the gradient updates.

In the subsequent training stages, we initialize the student model with the best FID checkpoint from the previous stage, and finetune with AdamW optimizer. For stage 2, we use a learning rate of 5e-6, a weight decay of 0.01, beta parameters $\left(0.9,0.999\right)$, batch size 256, 5 adapted-network updates per student update, and train for 35K iterations with 4 A100 GPUs. In total, PMDD processes approximately 60 million synthetic samples across the two training stages.

\subsection{ImageNet $64\times64$}

We distill a one-step student model from the EDM Pretrained model \cite{karras2022elucidating}, using edm-imagenet-64x64-cond-adm for class-conditional image generation. Training is performed with AdamW optimizer (learning rate of 2e-6, a weight decay 0.01, and beta parameters $\left(0.9,0.999\right)$), a batch size of 80 on 4 A100 GPUs for 400K iterations. Following the original implementation, we adopt a two-time update rule where the adapted network is updated 5 times per a student update.

Subsequent stages progressively extend the distilled student model to support multi-step sampling. Each stage is initialized from the best FID checkpoint of the previous stage and finetuned with AdamW. We use a learning rate of 8e-7, a weight decay of 0.01, beta parameters $\left(0.9,0.999\right)$, batch size 64, 5 adapted-network updates per student update, and train for 100K iterations with 4 A100 GPUs for subsequent stages. In total, the one-step model is trained on 32M synthetic samples, while the two-step model further uses 6.4M samples, i.e., $\approx$38M samples in total.

\subsection{SD v1.5 and PixArt$\alpha$-$1024$}

We distill a one-step student model from the SD v1.5 model \cite{rombach2022high} using prompts from LAION-Aesthetic 6.25+ dataset. Unlike DMD v2, our approach does not utilize real images for training the adversarial objective. Training is performed with AdamW optimizer (learning rate of 5e-6, a weight decay 0.01, and beta parameters $\left(0.9,0.999\right)$), a total batch size of 160 on 4 H100 GPUs for 350K iterations. Following prior work, we adopt a two-time update rule where the adapted network is updated 10 times per student update.

Subsequent stages progressively extend the distilled student model to support multi-step sampling. Each stage is initialized from the best FID checkpoint of the previous stage and finetuned with AdamW. We use a learning rate of 8e-7, a weight decay of 0.01, beta parameters $\left(0.9,0.999\right)$, batch size 20, 5 adapted-network updates per student update, unforget weight 0.3, and train for 16K iterations with 1 H100 GPU, which takes approximately 12 hours (0.32M synthetic samples). To accelerate convergence, we reduce the adapted network updates to 3 per student update, set the unforget weight to 0.005, and continue training for an additional 65K iterations on 4 H100 GPUs to obtain the 3-step PMDD model.

For the high-resolution setting, we apply PMDD to PixArt-$\alpha$, a Diffusion Transformer at $1024\times1024$, following the distillation and evaluation protocol of SANA-Sprint \cite{DBLP:conf/iccv/ChenXZYPCCHX25}. We keep the same progressive schedule and optimizer settings as above, distilling a one-step student and then extending it to 2- and 3-step students; the only change is the $1024\times1024$ backbone

\section{Evaluation Details}
\label{sec:Evaluation-Details}

For the CIFAR-10 and ImageNet $64\times64$ datasets, we generate 50,000 images every 1000 iterations, and calculate the FID metric based on EDM's evaluation code \cite{karras2022elucidating}. For the zero-shot COCO 2014 text-to-image generation, we closely follow the evaluation process as reported in DMD \cite{yin2024one}, in which 30,000 images are generated from prompts of the MS-COCO2014 validation set. The generated images are then resized to $256\times256$ before computing FID against the corresponding real validation set. We also report CLIP scores using the OpenCLIP-G model to assess text--image alignment. For the high-resolution setting, we evaluate on MJHQ-30k, reporting FID, CLIP, and GenEval following the protocol of SANA-Sprint \cite{DBLP:conf/iccv/ChenXZYPCCHX25}.

We aim at controlled comparisons: all distilled methods use a matched backbone, model size, resolution, and sampling budget, under the same training and evaluation protocol. When official numbers are unavailable and no public checkpoint exists, we reproduce results using the authors' official code. Where neither code nor checkpoints are released (\eg FS-SiD), we report only the values stated in the original paper and omit the method from experiments requiring inference (\eg qualitative and consistency comparisons).

\begin{figure}
\centering
\includegraphics[width=\columnwidth]{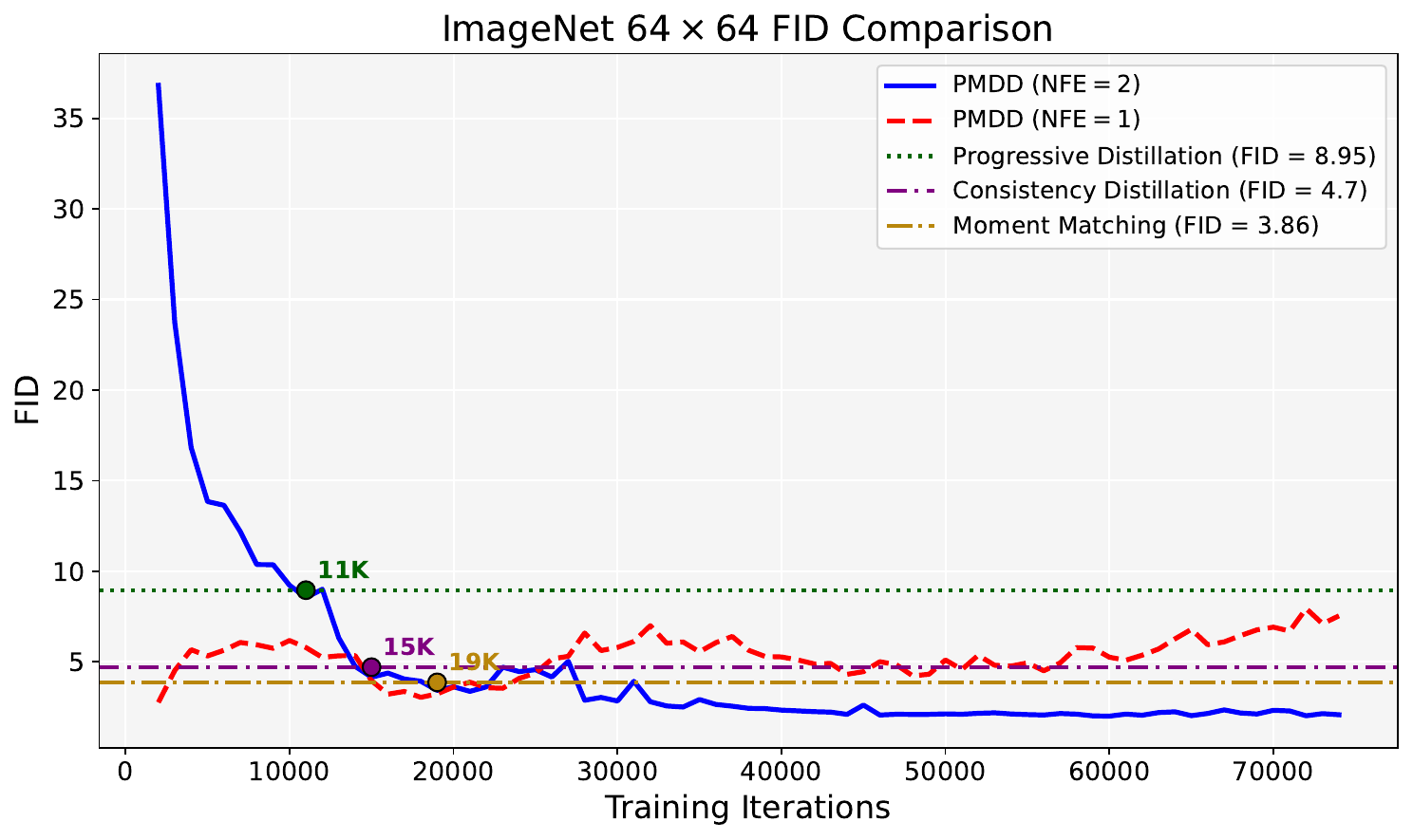}
\caption{FID on ImageNet $64\times64$ during the second (two-step) training stage, sampled with NFE$\,=2$ and NFE$\,=1$. Iterations count from the start of stage 2, which is initialized from the one-step student. Horizontal lines mark the 2-step baselines: Progressive Distillation (FID 8.95), Consistency Distillation (FID 4.70) and Moment Matching (FID 3.86); dots mark the first iteration at which PMDD (NFE$\,=2$) falls below each. \label{fig:Effect-of-unforget_cifar_imagenet-1}}
\end{figure}

\paragraph{Total Generated Synthetic Samples}

For class-conditional image generation on CIFAR-10, PMDD processes approximately 60 million synthetic samples across two training stages. On ImageNet 64$\times$64, training requires approximately 38 million synthetic samples across two stages, and the two-step model surpasses its EDM teacher.

For text-to-image generation, PMDD is trained with 57.6 million synthetic samples over three stages on at most four GPUs, compared with 256 TPU v5e chips for Multistep Moment Matching \cite{DBLP:journals/corr/abs-2406-04103}.

\paragraph{Convergence speed}

PMDD rapidly acquires the ability to generate high-fidelity images within two sampling steps. Later stages are cheap because they converge quickly: each starts from a trained student rather than from the teacher. On ImageNet64 (Figure~\ref{fig:Effect-of-unforget_cifar_imagenet-1}), the two-step FID passes Progressive Distillation after about 0.7M images (11K iterations, under 5 hours), Consistency Distillation after about 1.0M images (15K iterations, under 8 hours), and 2-step Moment Matching after about 1.2M images (19K iterations, under 10 hours), less than 4\% of the samples used in the first stage.

\section{Additional Results}

\subsection{CIFAR-10 Results}

Following the implementation and evaluation details, we train PMDD on class-conditional CIFAR-10 benchmark and conduct comparisons against other baselines. We report the results in Table \ref{tab:Results-on-cifar10}, where PMDD outperforms most other data-free and data-dependent distillation methods.

\begin{table}[h]
\centering
{\resizebox{0.90\columnwidth}{!}{%
\begin{tabular}{clcc}
\toprule
Type & Method & NFE $\left(\downarrow\right)$ & FID $\left(\downarrow\right)$ \\
\midrule
Teacher & VP-EDM \citep{karras2022elucidating} & 35 & 1.79 \\
\midrule
\multirow{8}{*}{One-step} & GET-Base \citep{yin2024one} & 1 & 6.25 \\
 & Meng et al. \citep{meng2023distillation} & 1 & 5.98 \\
 & 2-Rectified Flow$^{\dagger}$ \citep{liu2022flow} & 1 & 4.85 \\
 & DMD (w/o KL) \citep{yin2024one} & 1 & 3.82 \\
 & DMD \citep{yin2024one} & 1 & 2.66 \\
 & \textcolor{brown}{SwiftBrush}$^{\dagger}$ \citep{nguyen2024swiftbrush} & \textcolor{brown}{1} & \textcolor{brown}{4.46} \\
 & \textcolor{brown}{Diff-Instruct} \citep{DBLP:journals/corr/abs-2410-20898} & \textcolor{brown}{1} & \textcolor{brown}{4.19} \\
 & \textcolor{brown}{SiD} \citep{zhou2024score} & \textcolor{brown}{1} & \textbf{\textcolor{brown}{1.93}} \\
\midrule
\multirow{10}{*}{Multi-step} & \multirow{2}{*}{Prog. Distill.$^{\dagger}$ \citep{salimans2022progressive}} & 1 & 9.12 \\
 & & 2 & 4.51 \\
 & \multirow{2}{*}{TRACT$^{\dagger}$ \citep{berthelot2023tract}} & 1 & 3.78 \\
 & & 2 & 3.32 \\
 & \multirow{2}{*}{Cons. Distill.$^{\dagger}$ \citep{song2023consistency}} & 1 & 3.55 \\
 & & 2 & 2.93 \\
 & \textcolor{brown}{DF-DMDv2} \citep{yin2024improved} & \textcolor{brown}{1} & \textcolor{brown}{5.58} \\
 & & \textcolor{brown}{2} & \textcolor{brown}{3.27} \\
 & \multirow{2}{*}{\textcolor{brown}{PMDD (ours)}} & \textcolor{brown}{1} & \textcolor{brown}{2.52} \\
 & & \textcolor{brown}{2} & \underline{\textcolor{brown}{2.19}} \\
\bottomrule
\end{tabular}
}}
\caption{Results on CIFAR-10. PMDD uses $\mathcal{L}^{\text{stu}}_{\text{SiD}}$ (Appendix D.3). Data-free (DF) distillation methods are shown in \textcolor{brown}{brown} color and unconditional methods are indicated with $^{\dagger}$. The best and second best DF methods are highlighted in \textbf{bold} and \underline{underscored}, respectively.\label{tab:Results-on-cifar10}}
\end{table}

\subsection{Precision and Recall on SDv1.5}

FID conflates fidelity and diversity, so we additionally report improved precision/recall on SDv1.5 (Table~\ref{tab:precision_recall}). PMDD outperforms data-free DMDv2 at matched steps, indicating the FID gains reflect a genuine distribution match rather than a fidelity-diversity trade-off. We compare against DMDv2 because FS-SiD releases neither code nor checkpoints.

\begin{table}
\centering
\small
\begin{tabular}{lcc}
\toprule
Method & Precision $\left(\uparrow\right)$ & Recall $\left(\uparrow\right)$ \\
\midrule
\multicolumn{3}{l}{\textit{2-step}} \\
DMDv2 & 0.59 & 0.46 \\
PMDD & \textbf{0.62} & \textbf{0.53} \\
\midrule
\multicolumn{3}{l}{\textit{3-step}} \\
DMDv2 & 0.62 & 0.46 \\
PMDD & \textbf{0.64} & \textbf{0.52} \\
\bottomrule
\end{tabular}
\caption{Precision and recall of data-free DMDv2 and PMDD on SD v1.5. \label{tab:precision_recall}}
\end{table}

\subsection{Student Loss}

We propose a progressive training framework that aligns the student distribution with the teacher distribution. As an orthogonal investigation, we study the integration of the score identity loss $\mathcal{L}^{\text{stu}}_{\text{FS-SiD}}\left(\phi\right)$ to examine whether it improves student performance.

Table \ref{tab:unforget_loss_2step_cifar_imagenet-2} summarizes the main results. The findings show that class-conditional generation performance is mainly driven by $\mathcal{L}^{\text{stu}}_{\text{FS-SiD}}\left(\phi\right)$ under 2-step inference, achieving better performance than the per-step formulation. However, $\ensuremath{\mathcal{L}^{i}_{\text{SiD}}(\phi)}$ also fails to scale to more complex settings such as text-to-image generation. Due to the lack of publicly available implementation details for FS-SiD, a deeper investigation is left for future work.

In contrast, $\mathcal{L}^{\text{stu}}_{\text{DM}}\left(\phi\right)$ shows consistent improvements as the number of sampling steps increases across datasets, ranging from CIFAR-10 to large-scale text-to-image generation with Stable Diffusion v1.5 \cite{rombach2022high}.

\begin{table}
\centering
\small
\begin{tabular}{ccc}
\toprule
NFE & $\mathcal{L}^{\text{stu}}_{\text{DM}}$ & $\mathcal{L}^{\text{stu}}_{\text{SiD}}$ \\
\midrule
1 & 3.49 & 2.52 \\
2 & 2.91 & \textbf{2.19} \\
\bottomrule
\end{tabular}
\caption{Effect of the student loss on CIFAR-10 (FID). \label{tab:unforget_loss_2step_cifar_imagenet-2}}
\end{table}

\subsection{Unforgetting Loss Weight $\alpha$}

We further examine the influence of varying the unforget loss weight $\alpha$ on CIFAR-10 in Table \ref{tab:unforget_weight_2step_sd-1}. Across a broad range of values, the overall performance remains stable, indicating that the proposed framework is not sensitive to this hyperparameter. Larger weights slightly reduce catastrophic forgetting across previously learned steps but may slow convergence and mildly affect the performance of the current step.

\begin{table}
\centering
\small
\begin{tabular}{lcc}
\toprule
 & FID (NFE$\,=2$) & FID (NFE$\,=1$) \\
\midrule
$\alpha=0.1$ & \textbf{2.19} & 7.84 \\
$\alpha=1.0$ & 2.21 & 6.78 \\
$\alpha=3.0$ & 2.21 & \textbf{4.91} \\
\bottomrule
\end{tabular}
\caption{Effect of unforget weight $\alpha$ on 2-step PMDD (CIFAR-10). \label{tab:unforget_weight_2step_sd-1}}
\end{table}

In practice, we observe marginal differences across unforgetting loss weight settings $\alpha$ on CIFAR-10 (Table \ref{tab:unforget_weight_2step_sd-1}). A default value of $\alpha=0.1$ provides a good balance between limiting forgetting and maintaining efficient training. On SD v1.5, we use $\alpha=0.3$ when stage 2 is introduced and a smaller $\alpha=0.005$ in stage 3, placing more emphasis on optimizing the current step, consistent with Figure \ref{fig:Effect-of-unforget_text_2_img}. These results support our hypothesis that overly large unforgetting weights over-constrain the student, whereas moderate weighting preserves sufficient flexibility for the model to better match the current-step distribution.

\begin{figure*}
\centering
\includegraphics[width=0.90\linewidth]{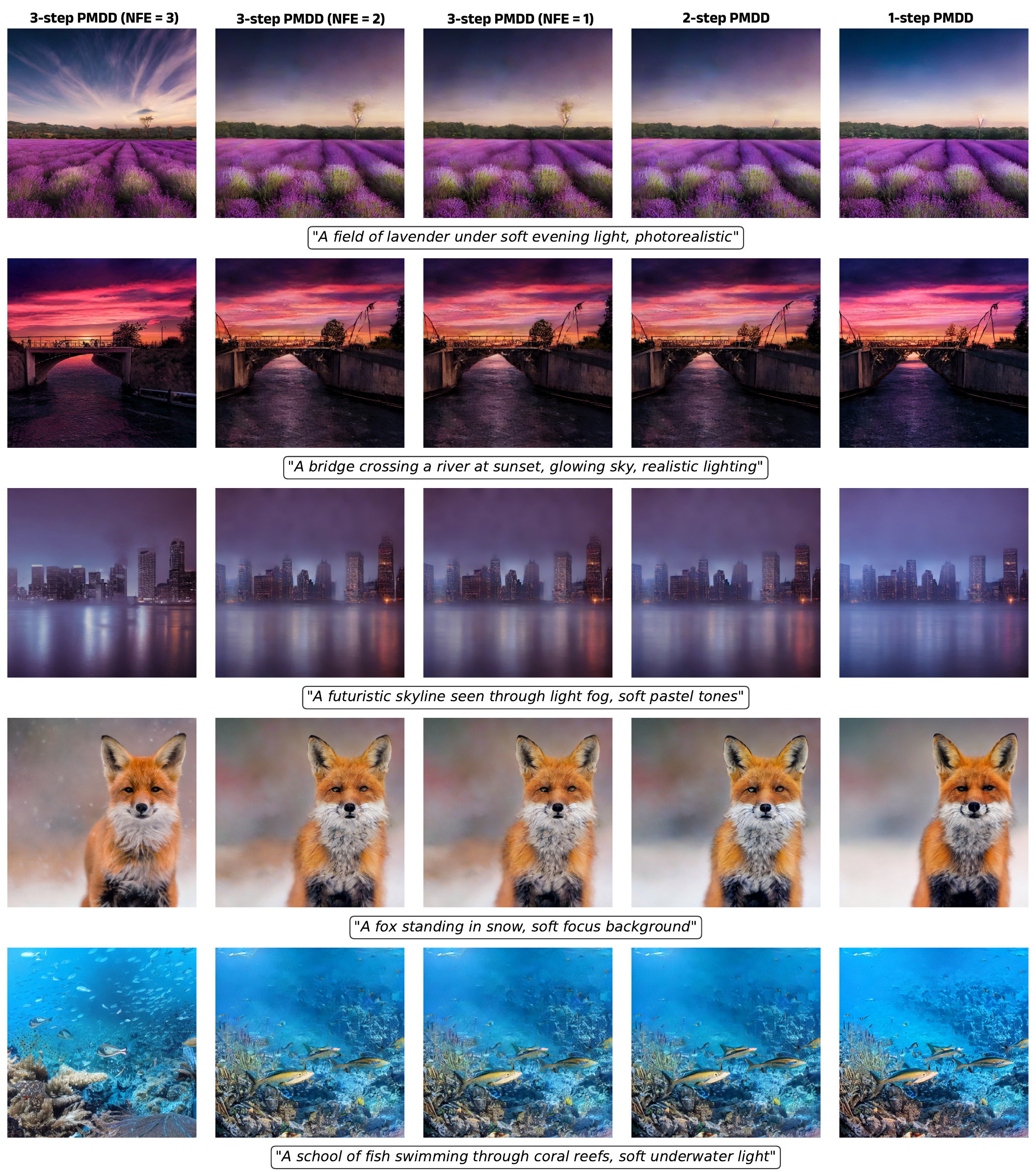}
\caption{Additional qualitative comparison of 3-step PMDD outputs (NFE = 3, CFG = 1.0) with previously generated 1-step and 2-step reference images. \label{fig:Uncurated-samples_unforget}}
\end{figure*}

Figure \ref{fig:Uncurated-samples_unforget} shows a qualitative analysis of how the unforget weight influences consistency on 3-step PMDD with previously generated PMDD samples (originally produced using 1-step and 2-step models). The results suggest that the framework remains robust to the choice of unforgetting weight. Even with a small value ($\alpha=0.01$), the 3-step model preserves the global structure of outputs from earlier 1-step and 2-step models (Ref), with only minor variations in fine details. The 3-step model also demonstrates an improvement capability: relative to its reference images, it can adjust specific details to better follow the prompt, such as ``school of fish'' in Figure \ref{fig:Uncurated-samples_unforget}.

\subsection{Effect of Previous-Step Sampling}

\begin{figure*}[t]
\centering
\includegraphics[width=0.95\linewidth]{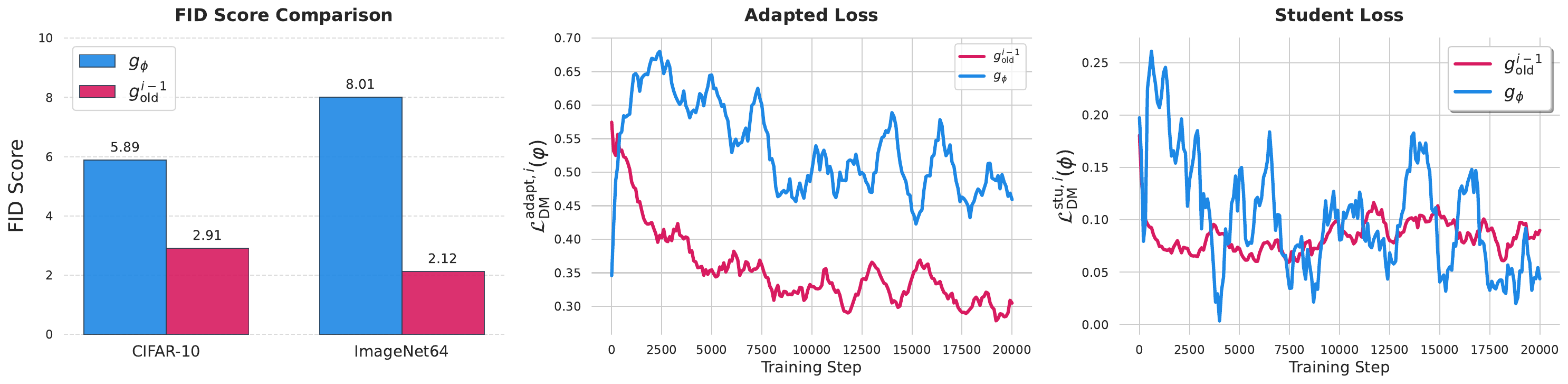}
\caption{Ablation study of using $g^{i-1}_{\text{old}}$ for sampling $x_{t_{i}}$ (2-step, $\mathcal{L}^{\text{stu}}_{\text{DM}}$). Left: FID on CIFAR-10 and ImageNet64. Middle/right: adapted-network and student losses on ImageNet64.\label{fig:previous_step_sampling_app}}
\end{figure*}

Sampling $x_{t_{i}}$ from $g^{i-1}_{\text{old}}$ instead of the evolving student lowers FID from 5.89 to 2.91 on CIFAR-10 and from 8.01 to 2.12 on ImageNet64 (Figure~\ref{fig:previous_step_sampling_app}). The adapted loss remains consistently lower and decreases smoothly throughout training, whereas the student loss shows noticeably lower variability.

\subsection{Stage-conditioned Adapted Network}
\label{app:stage_conditioned}

\begin{table}[h]
\centering
\small
\setlength{\tabcolsep}{4pt}
\begin{tabular}{lccc}
\toprule
Method & Adapted network & CIFAR-10 & ImageNet64 \\
\midrule
DF-DMDv2 & shared & 3.27 & 2.49 \\
DF-DMDv2 & stage-conditioned & 3.22 & 2.44 \\
PMDD & per stage & \textbf{2.19} & \textbf{1.99} \\
\bottomrule
\end{tabular}
\caption{Stage conditioning of the adapted network (2-step, FID). DF-DMDv2 rows use $\mathcal{L}^{\text{stu}}_{\text{DM}}$; PMDD results are from Table~\ref{fig:main_table}(b) and Table~\ref{tab:Results-on-cifar10}. \label{tab:stage_conditioned}}
\end{table}

We condition the shared adapted network of DF-DMDv2 on the stage by adding a learned embedding of the stage index $i$ to its timestep embedding; all other settings follow DF-DMDv2. Stage conditioning changes FID by only 0.05 on both datasets (3.27 $\rightarrow$ 3.22 on CIFAR-10, 2.49 $\rightarrow$ 2.44 on ImageNet64, Table~\ref{tab:stage_conditioned}). With the same $\mathcal{L}^{\text{stu}}_{\text{DM}}$ objective, PMDD reaches 2.91 on CIFAR-10 (Table~\ref{tab:unforget_loss_2step_cifar_imagenet-2}) and 2.12 on ImageNet64 (Figure~\ref{fig:previous_step_sampling_app}). The gain therefore comes from the progressive design rather than from conditioning the adapted network.

\section{Extended Related Work}

In this work, we focus on distributional matching diffusion distillation, which can be broadly grouped into two categories: either one-step or multi-step. Single-step generation focuses on aligning the distributions of the student and teacher across different time steps. SwiftBrush adapts 3D distribution matching techniques from Score Distillation Sampling \cite{poole2022dreamfusion} and Variational Score Distillation \cite{wang2024prolificdreamer} to 2D image synthesis by replacing the 3D NeRF rendering component with a 2D text-to-image generator. To further improve image fidelity, DMD \cite{yin2024one} augments the framework with an additional regression loss on synthetic data, while other methods rely on adversarial objectives during distillation \cite{DBLP:conf/iclr/Wang0HW025,DBLP:conf/cvpr/0003ZXH24}. Score Identity Distillation \cite{zhou2024score} generalizes this idea by replacing the reverse KL-Divergence used in original work with Fisher Divergence, featuring DMD as its special case while remaining data-free.

Several methods have been designed to extend distillation from one-step to multistep settings. Multistep Moment Matching \cite{DBLP:journals/corr/abs-2406-04103} performs first-moment alignment and attains state-of-the-art results; however, it operates in pixel space rather than latent space, leading to substantial model size and computational overhead. DMDv2 \cite{yin2024improved} and FS-SiD \cite{DBLP:journals/corr/abs-2505-12674} extend one-step distribution matching to multiple steps with a single shared adapted network; TDM \cite{DBLP:conf/iccv/LuoHSCT25} combines both distribution and trajectory matching approaches, and Decoupled DMD \cite{DBLP:journals/corr/abs-2511-22677} improves few-step text-to-image distillation through decoupled training objectives. All retain a shared score estimator across steps; PMDD instead decouples score estimation per stage via progressive mechanism, yielding cross-step consistency and a single model for all budgets.

\section{Prompts for Figures}
\label{subsec:additional_qualitative-1}

For visualizations, we use prompts from the LAION-Aesthetic 6.25+ dataset; a number of them were also used for the qualitative results in prior work, enabling a direct side-by-side comparison \cite{yin2024one,yin2024improved} . The list of prompts is:

\begin{itemize}
\item Create an image that depicts a majestic kingdom with towering castles, lush gardens, and vibrant colors. show the bustling streets filled with lively townsfolk and capture the enchanting atmosphere of the realm.
\item A dog of the german shepherd breed, wearing a space suit, in a post-pocalyptic world, among debris, rubble, dramatic colors, cinematic lighting, vivid colors, sparkling colors, full colors vector.
\item A surreal city floating in the clouds, impossible architecture, vivid colors, ultra-detailed.
\item The giant magical deer god of the forest, sniffing flowers on the forest floor. Fireflies evereywhere. A spring of water. Long moss hanging from the tree branches. Moonlight. Photorealism, cinematic shot, cinematic lighting, National Geographic, analagous colors, Award-winning photograph.
\item Border collie surfing a small wave, with a mountain on background.
\item Futuristic lighthouse, flash light, hyper realistic, epic composition, cinematic.
\item A lake in autumn with golden trees and leaves, photorealistic.
\item A realistic portrait of an older man with light-blonde hair and a full white beard, facing forward with a calm expression. Warm lighting, dark background, oil-painting style.
\item A waterfall in a tropical jungle, high resolution, 4k.
\item A pride of lions resting in the savannah under golden sunset light, detailed, 4k.
\item A crowded Tokyo street during cherry blossom season, highly detailed, realistic lighting.
\item Macro photography of a realistic and natural ladybug.
\item An intricate steampunk machine with gears and pipes, highly detailed, dramatic lighting.
\end{itemize}

\section{Additional Qualitative Samples}
\label{subsec:additional_qualitative_final}

\begin{figure*}[H]
\centering
\includegraphics[width=0.98\linewidth]{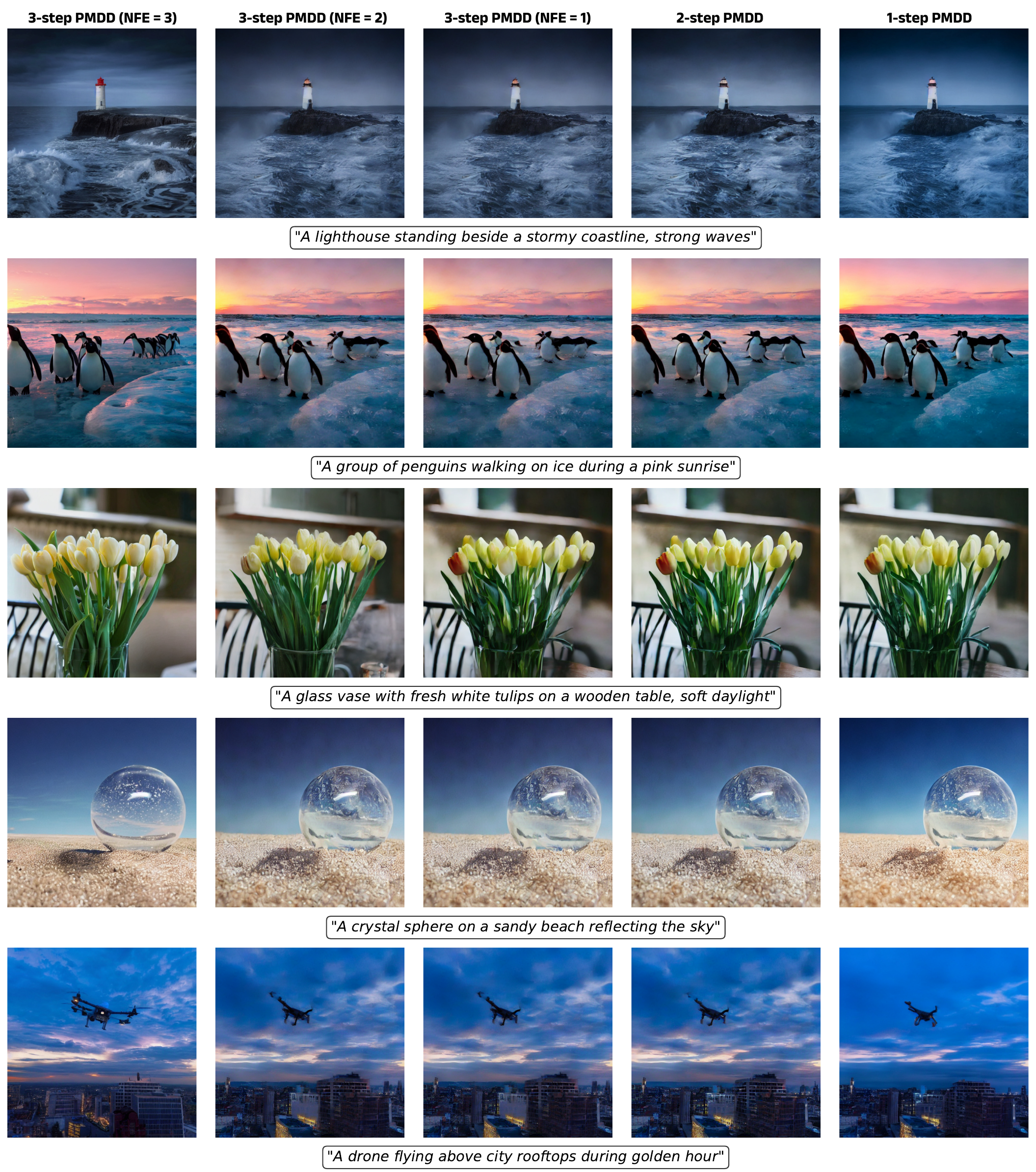}
\caption{Qualitative comparison of 3-step PMDD outputs (NFE = 3, CFG = 1.0) with previously generated 1-step and 2-step reference images. \label{fig:Uncurated-samples-4.5-1-1}}
\end{figure*}

\clearpage

\begin{figure*}
\centering
\includegraphics[width=1.00\linewidth]{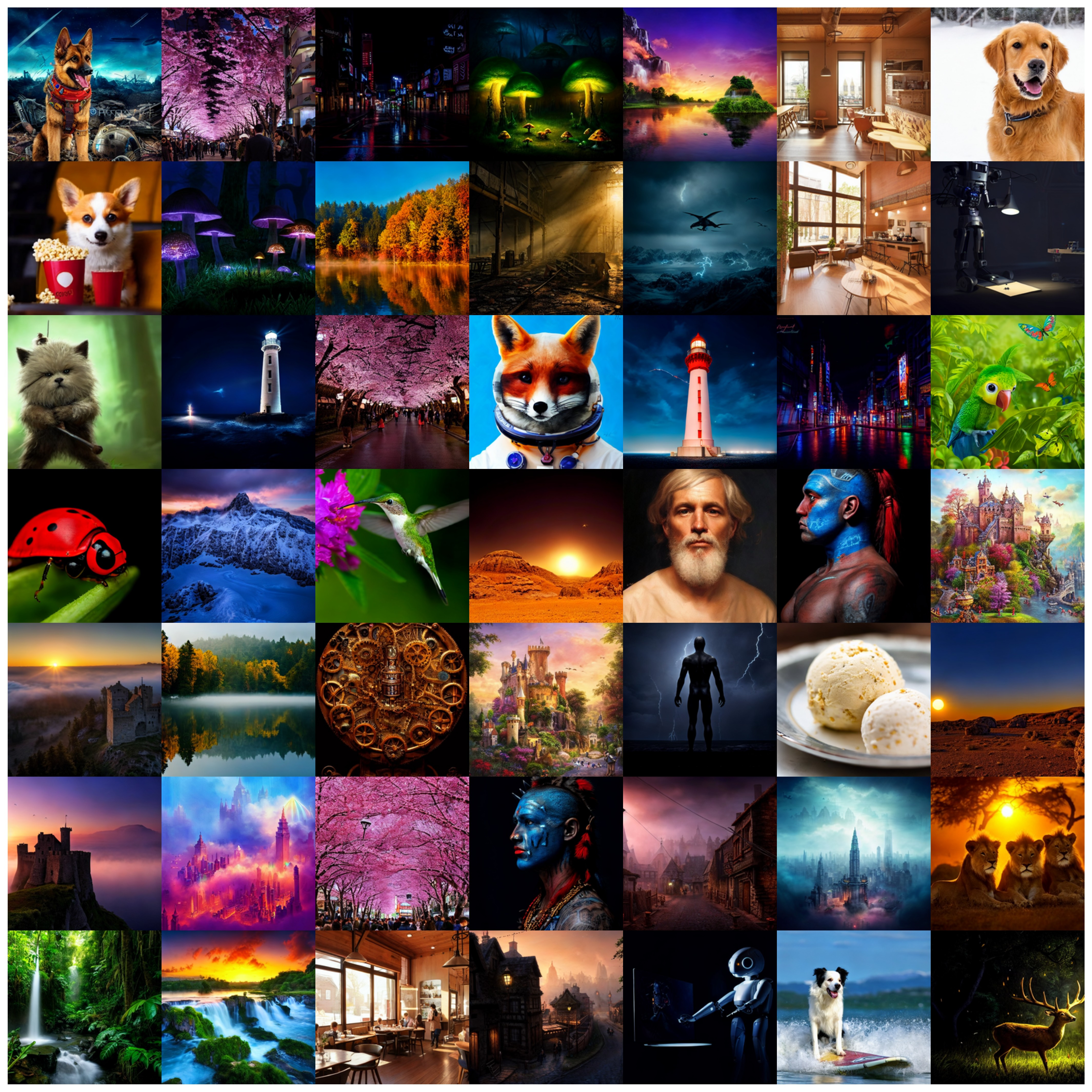}
\caption{Uncurated samples of our text-to-image model\label{fig:Uncurated-samples-4.5}}
\end{figure*}

\clearpage

\begin{figure*}
\centering
\includegraphics[width=1.00\linewidth]{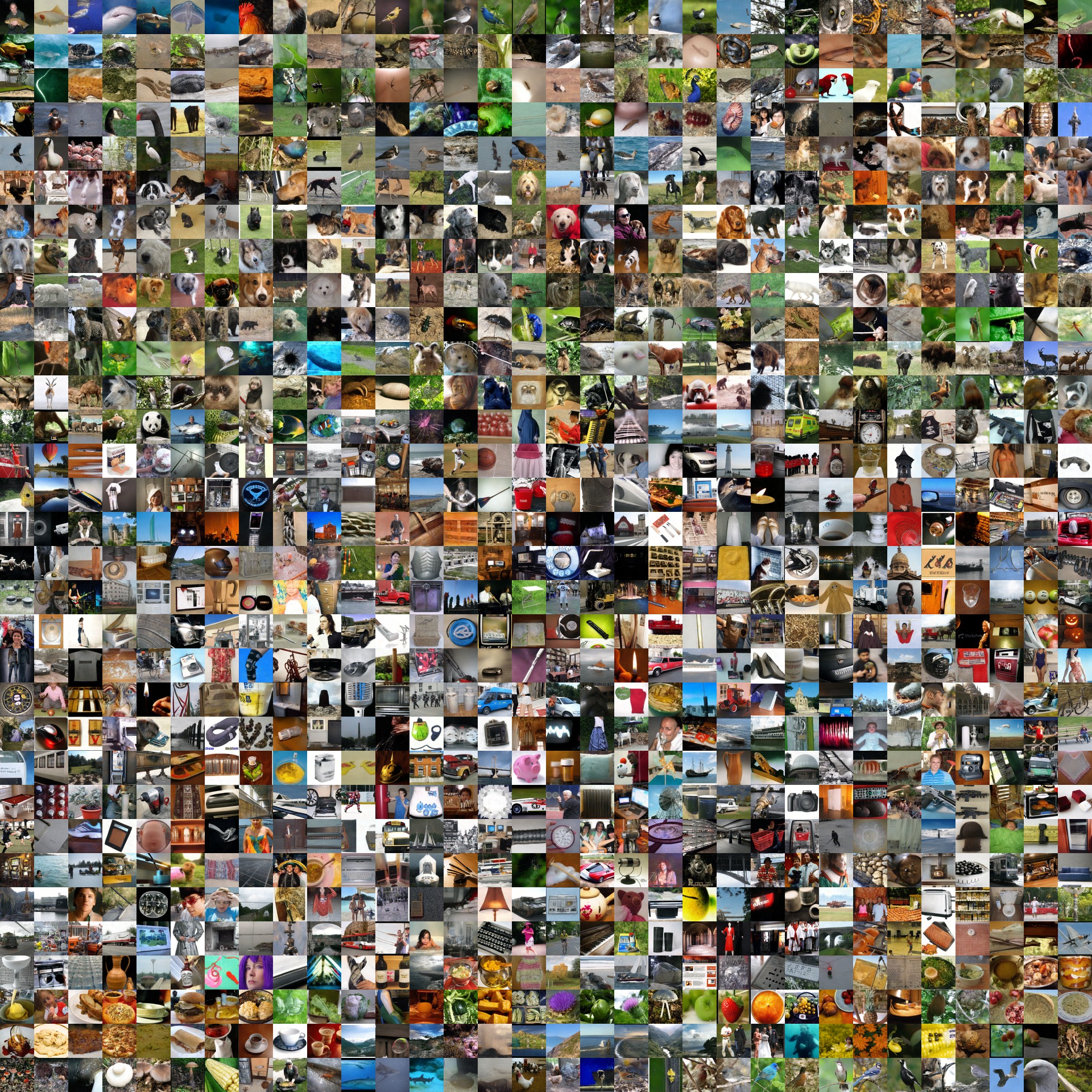}
\caption{Samples from our 1-step student generator on ImageNet (FID=2.58).}
\end{figure*}

\clearpage

\begin{figure*}
\centering
\includegraphics[width=1.00\linewidth]{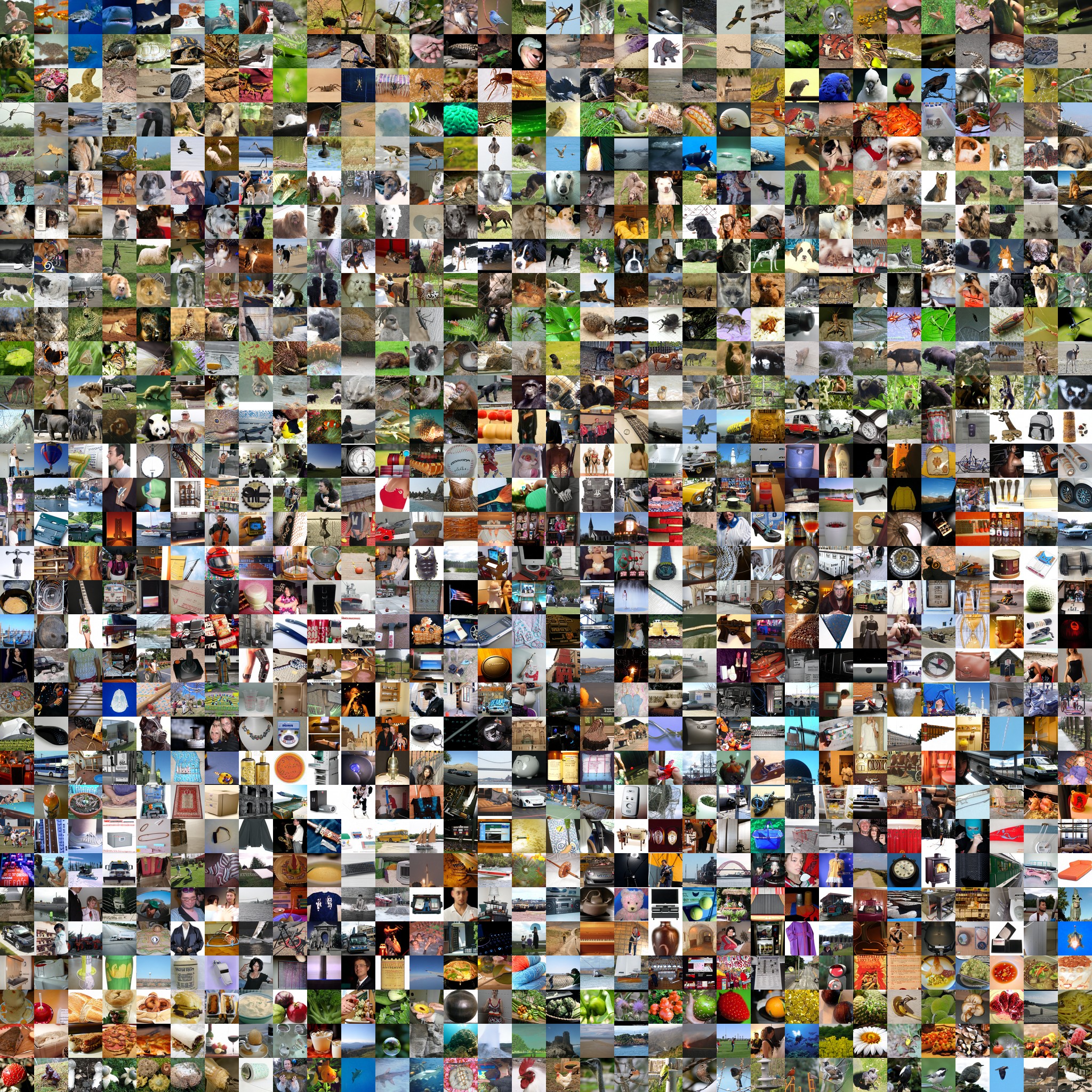}
\caption{Samples from our 2-step student generator on ImageNet (FID=1.99).}
\end{figure*}

\clearpage

\section{Algorithm}

\begin{algorithm*}[t]
\caption{PMDD: Progressive Multistep Distillation}
\textbf{Input:} teacher $x_{\theta}$; adapted net $x_{\varphi}$; student $g_{\phi}$;
frozen previous student $g_{\text{old}}$; steps $N$; schedule $0<t_N<\dots<t_1=T$;
unforget weight $\alpha$.
\begin{algorithmic}[1]
\For{$i = 1,\ldots,N$} \Comment{stage-wise training}
    \If{$i = 1$}
        \State $g_{\phi}^{1},\,x_{\varphi}^{1} \gets x_{\theta}$ \Comment{init from teacher}
    \Else
        \State $g_{\phi}^{i},\,x_{\varphi}^{i} \gets g_{\phi}^{i-1},\,x_{\varphi}^{i-1}$ \Comment{init from previous stage}
        \State $g_{\text{old}}^{i-1} \gets g_{\phi}^{i-1}$ \Comment{frozen reference}
    \EndIf
    \State
    \State \textit{// recursive sampling of $x_{t_i}$ via the frozen student}
    \State $x_{t_1} \gets x_T,\quad x_T \sim \mathcal{N}(0,\mathrm{I})$
    \For{$k = 1,\ldots,i-1$}
        \State $\bar{x}_{0|\phi,t_k}=\text{sg}\big(g_{\text{old}}^{i-1}(x_{t_k},t_k)\big)$
        \State $x_{t_{k+1}} = a_{t_{k+1}}\bar{x}_{0|\phi,t_k} + \sigma_{t_{k+1}}\epsilon$
    \EndFor
    \State $x_{0|\phi,t_i} \gets g_{\phi}^{i}(x_{t_i},t_i)$
    \State
    \State \textit{// update adapted network}
    \State $t \sim \mathcal{U}(1,T),\quad x_t \gets a_t x_{0|\phi,t_i}+\sigma_t\epsilon$
    \State $\Loss^{\text{adapt},i}_{\text{DM}}(\varphi) \gets \Expect\big[w_x(t)\,\|x_{\varphi}(x_t,t)-x_{0|\phi,t_i}\|_2^2\big]$
    \State update $x_{\varphi}^{i}$
    \State
    \State \textit{// update student}
    \State $\Loss^{\text{stu},i}_{\text{DM}}(\phi) \gets \Expect\big[w_x(t)\,\|x_{\varphi}(x_t,t)-x_{\theta}(x_t,t)\|_2^2\big]$
    \If{$i \neq 1$}
        \State $\Loss^{\text{stu},i}_{\text{unf}}(\phi) \gets \sum_{k=1}^{i-1}\Expect\big[\|g_{\phi}^{i}(x_{t_k},t_k)-g_{\text{old}}^{i-1}(x_{t_k},t_k)\|_2^2\big]$
    \EndIf
    \State $\Loss^{\text{stu},i}(\phi) \gets \Loss^{\text{stu},i}_{\text{DM}}(\phi) + \alpha\,\Loss^{\text{stu},i}_{\text{unf}}(\phi)$
    \State update $g_{\phi}^{i}$
\EndFor
\State \textbf{return} $g_{\phi}^{N}$
\end{algorithmic}
\end{algorithm*}

\clearpage